\documentclass[runningheads]{llncs}

\usepackage[mobile]{eccv}

\usepackage{eccvabbrv}

\usepackage{graphicx}
\usepackage{booktabs}
\usepackage{xspace}
\usepackage{adjustbox}
\usepackage{makecell}
\usepackage{placeins}
\usepackage[table]{xcolor}
\usepackage{bibunits}
\definecolor{lightblue}{RGB}{220,235,255}

\usepackage[accsupp]{axessibility}  %

\newcommand{\method}{MIRAGE\xspace}
\newcommand{\methodlong}{Multi-Instance Regional Alignment via Guided Editing\xspace}
\newcommand{\bench}{MIRA-Bench\xspace}

\usepackage{hyperref}

\usepackage{orcidlink}
\usepackage[utf8]{inputenc}
\usepackage[T1]{fontenc}
\usepackage{amsmath, amssymb}
\usepackage{caption}
\usepackage{subcaption}
\usepackage{xcolor}
\usepackage{float}
\usepackage{textcomp}
\usepackage{tabularx}
\usepackage{multirow}
\usepackage{colortbl}
\usepackage{diagbox}
\usepackage{adjustbox}
\usepackage{array}
\usepackage{dblfloatfix}
\usepackage{makecell}

\usepackage{arydshln}   
\usepackage{graphicx}  

\begin{document}

\title{MIRAGE: Benchmarking and Aligning Multi-Instance Image Editing}

\author{Ziqian Liu%
\and
Stephan Alaniz%
}

\authorrunning{Z.~Liu and S.~Alaniz}

\institute{LTCI, Télécom Paris, Institut Polytechnique de Paris, France
}

\maketitle

\begin{abstract}
Instruction-guided image editing has seen remarkable progress with models like FLUX.2 and Qwen-Image-Edit, yet they still struggle with complex scenarios with multiple similar instances each requiring individual edits. We observe that state-of-the-art models suffer from severe over-editing and spatial misalignment when faced with multiple identical instances and composite instructions. To this end, we introduce a comprehensive benchmark specifically designed to evaluate fine-grained consistency in multi-instance and multi-instruction settings.
To address the failures of existing methods observed in our benchmark, we propose \methodlong (\method), a training-free framework that enables precise, localized editing. By leveraging a vision-language model to parse complex instructions into regional subsets, \method employs a multi-branch parallel denoising strategy. This approach injects latent representations of target regions into the global representation space while maintaining background integrity through a reference trajectory. Extensive evaluations on \bench and RefEdit-Bench demonstrate that our framework significantly outperforms existing methods in achieving precise instance-level modifications while preserving background consistency. Our benchmark and code are available at \url{https://github.com/ZiqianLiu666/MIRAGE}.

\end{abstract}

\section{Introduction}

The field of visual content creation has been revolutionized by the emergence of diffusion models~\cite{stablediffusion, diffusion, sdxl}, which now achieve unprecedented fidelity in image synthesis. This success has paved the way for instruction-guided image editing~\cite{ip2p, magicbrush}, where users can modify complex visual scenes using only natural language. While these models excel at global transformations and single object adjustments, current models still struggle with handling multi-instance scenarios where multiple similar objects require distinct, independent modifications.

\begin{figure}
\centering
\includegraphics[width=\columnwidth]{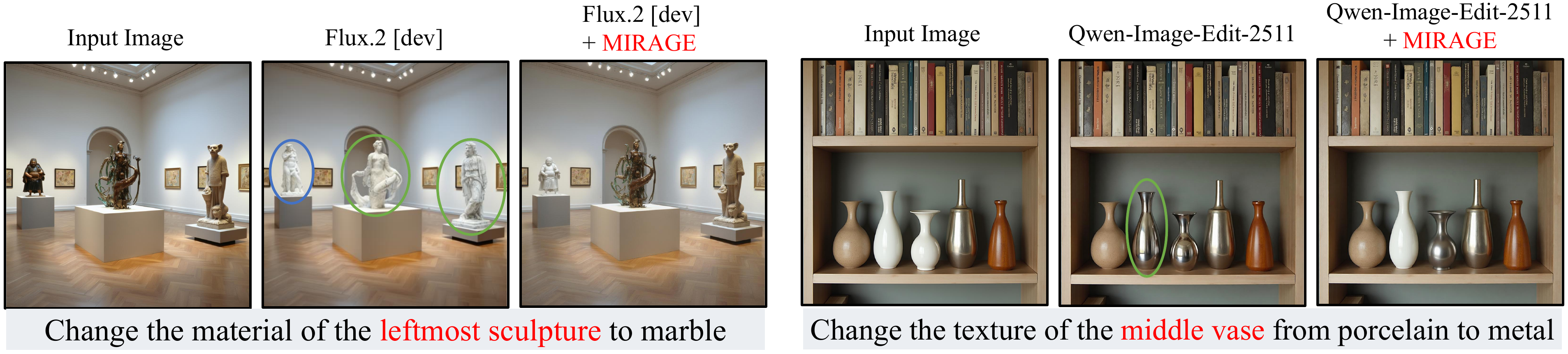}
\caption{\textbf{Limitations of current state-of-the-art image editing models.} Although these models successfully follow the editing instructions, noticeable limitations remain. (i) blue circles indicate failures in preserving local consistency within the target regions; (ii) green circles highlight unintended modifications in the background.
} 
\label{fig:teaser}
\end{figure}

In these complex settings, even state-of-the-art (SOTA) models like FLUX.2~\cite{flux.2} and Qwen-Image-Edit~\cite{qwen-edit} exhibit spatial misalignment and over-editing. For example, as shown in Fig.~\ref{fig:teaser}, when given the instruction “Change the material of the leftmost sculpture to marble” in a scene containing three sculptures, FLUX.2 changes the material of \emph{all} sculptures in the image.

Existing benchmarks lack the density and complexity required to diagnose these fine-grained errors. 
To better measure progress in such challenging settings,
we introduce \bench, a comprehensive benchmark specifically designed to evaluate fine-grained consistency in multi-instance and multi-instruction scenarios. While previous benchmarks like MagicBrush ~\cite{magicbrush} or RefEdit-Bench ~\cite{pathiraja2025refedit} focus on single-object edits or referring expressions, \bench is the first to systematically test models against images containing three to five similar instances with five simultaneous, compositional edit instructions. This benchmark provides the community with a rigorous diagnostic tool for assessing model performance in complex editing tasks.

To address the challenge of accurately editing images with fine-grained instructions on the model side, we propose \methodlong (\method), a training-free framework designed for precise, localized image editing. Unlike prior approaches that require fine-tuning or rely on attention maps, \method leverages the inherent reasoning of Vision-Language Models (VLMs)~\cite{qwen3-vl} to decompose complex, global instructions into regional sub-tasks. By introducing a multi-branch parallel denoising strategy, \method enables the independent editing of specific instances while enforcing background integrity through a reference trajectory. This allows the model to perform high-fidelity edits on specific targets without disturbing the rest of the scene. As shown in Fig.~\ref{fig:teaser}, \method resolves the over-editing issues of FLUX.2 and Qwen-Image-Edit resulting in more reliable editing results.

In summary, our contributions are as follows: i) We introduce \bench, a novel benchmark focused on multi-instance scenes and compositional instructions, filling an important gap in image editing evaluation; ii) We propose \method, a training-free, multi-branch inference framework that improves the precision of mainstream diffusion models without additional training or computational overhead; iii) Through extensive experiments, we demonstrate that integrating \method into SOTA models yields significant gains in prompt following and consistency on both \bench and the existing RefEdit benchmark, effectively resolving persistent issues of over-editing and spatial misalignment.

\section{Related Work}

\textbf{Text-Guided Image Editing}. As diffusion models have made significant progress in image generation~\cite{stablediffusion,sdxl,dalle2}, they exhibit superior detail fidelity and local controllability compared to GAN-based~\cite{stylegan} methods, and have increasingly become the dominant paradigm for image editing. E4C~\cite{E4C} and FlexiEdit~\cite{FlexEdit} invert images into the latent space and incorporate attention constraints to improve structural consistency. However, such methods typically rely on correspondences between source and target texts, which introduces inconvenience in practical usage. Subsequently, InstructPix2Pix~\cite{ip2p} performed conditional fine-tuning to enable Stable Diffusion to learn a mapping from $(\mathrm{X_{src}}, \mathrm{Instruction})$ to $\mathrm{X_{trg}}$. Although this approach improves both performance and usability over prior methods, it is prone to over-editing~\cite{wys,foi,zone}, motivating a large body of follow-up work that explores improvements from the perspectives of attention control~\cite{foi,zone}, spatial conditioning~\cite{gip2p}, and inference strategies~\cite{earlytimestep,wys}. On the other hand, recent studies indicate that replacing the conventional U-Net architecture with Transformers leads to stronger generative performance~\cite{dit,dit4edit,lazy-dit}. Representative SOTA models, such as Qwen-Image-Edit~\cite{qwen-edit} and FLUX~\cite{flux.1,flux.2}, achieve significant gains on multiple generation and editing benchmarks. However, in scenarios involving multiple similar instances and compositional instructions, we observe that SOTA models~\cite{flux.2,qwen-edit}, while broadly following the editing intent, frequently exhibit over-editing, fail to preserve fine-grained detail consistency in target regions, and introduce unintended changes to non-target background areas. 
With \method, we effectively mitigate these issues with a training-free method without computational overhead.

\noindent\textbf{Localization of Regions of Interest.} Accurate localization of target regions is critical for consistency of detailed textures and for avoiding over-editing. Even for SOTA generative models~\cite{qwen-edit,flux.2,gpt-image}, reliably identifying the target object in complex scenes remains challenging. Among existing approaches, FOI~\cite{foi}, LOV~\cite{lov}, and LIME~\cite{lime} extract spatial masks by analyzing attention maps or feature attributions, while DiffEdit~\cite{DiffEdit} and WYS~\cite{wys} infer editing regions based on differences between predicted noise signals. These methods rely on model-internal signals, whose generalization tends to degrade in complex settings such as multi-instance scenes or when handling semantically complex instructions. In contrast, GIP2P~\cite{gip2p}, InstructEdit~\cite{instructedit}, and OIR~\cite{oir} leverage external models~\cite{groundingdino,sam} to localize target regions, though their performance is sensitive to the fine-grained accuracy of the external masks. 
RIE~\cite{referringedit} disentangles referring expressions from editing instructions via a self-attention mechanism and relies on an additional trained module to predict target masks for local edits. However, this design introduces additional training cost and computational overhead.
Our work falls into the category of external priors, and we employ a vision–language model (VLM)~\cite{qwen3-vl} to parse instructions and ground target references, enabling more precise localization of regions of interest.

\begin{table}[t]
\centering
\caption{\textbf{Comparison of image editing benchmarks.} 
\bench is the first benchmark to focus on multi-instance editing with 3-5 similar instances and 5 compositional edit instructions per image.}
\label{tab:benchmark}
\scriptsize
\renewcommand{\arraystretch}{1.12}
\setlength{\tabcolsep}{3.5pt}

\begin{tabular}{lcccccc}
\toprule
Dataset 
& \makecell[c]{Num.\\Image}
& \makecell[c]{Multi-\\Instance}
& \makecell[c]{Avg.\\Instances}
& \makecell[c]{Compositional\\Instructions}
& \makecell[c]{Avg. Instructions\\ / Image}
& Referable \\
\midrule
MagicBrush~\cite{magicbrush} & 535 & $\times$ & 1 & $\times$ & 2 & $\times$ \\
HIVE~\cite{hive}             & 200 & $\times$ & 1 & $\times$ & 5 & $\times$ \\
OmniEdit~\cite{omniEdit}     & 62  & $\times$ & 1 & $\times$ & 7 & $\times$ \\
ImgEdit~\cite{imgEdit}       & 811 & Partial  & 2 & Partial  & 4 & $\checkmark$ \\
RefEdit-Bench~\cite{pathiraja2025refedit} & 200 & Partial & 3 & Partial & 1 & $\checkmark$ \\
\midrule
\textbf{\bench} & \textbf{100} & $\checkmark$ & \textbf{4} & $\checkmark$ & \textbf{5} & $\checkmark$ \\
\bottomrule
\end{tabular}

\end{table}

\noindent\textbf{Instruction-Guided Image Editing Benchmarks.} Table~\ref{tab:benchmark} compares representative benchmarks designed for image editing tasks. Early benchmarks such as MagicBrush~\cite{magicbrush}, HIVE~\cite{hive}, and OmniEdit~\cite{omniEdit} primarily focus on relatively simple scenarios, where editing typically targets a single salient object or a local region in the image. Such settings rarely capture complex real-world scenes containing multiple entities, which limits their ability to evaluate model performance on multi-instance editing tasks.
More recent benchmarks, including ImgEdit~\cite{imgEdit} and RefEdit-Bench~\cite{pathiraja2025refedit}, expand the diversity of editing types and evaluation scope. However, only a limited portion of their samples involve multi-instance scenarios, and most evaluations still focus on single-instruction editing. As a result, these benchmarks cannot fully reflect model performance under complex compositional editing settings.
In contrast, our proposed \bench is specifically designed for multi-instance editing scenarios.

\section{\bench}

\begin{figure}[t]
\centering
\includegraphics[width=\columnwidth]{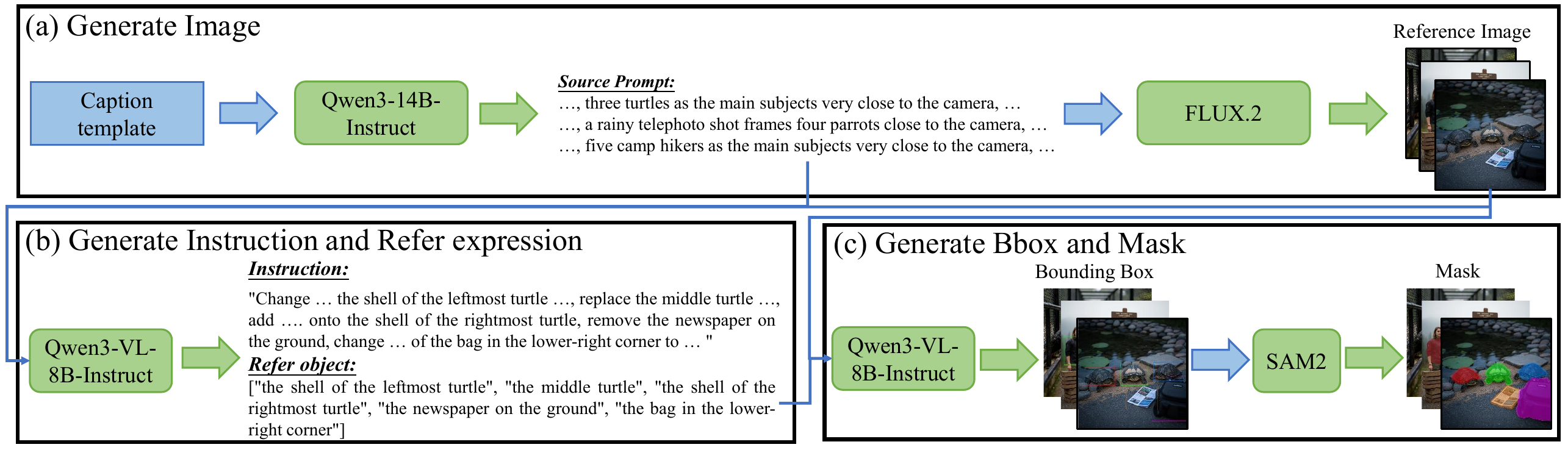}
\caption{\textbf{\bench construction.}
(a) Images are generated from source prompts via FLUX.2~\cite{flux.2}. (b) A VLM extracts editing instructions and referring expressions. (c) VLM-predicted boxes are refined by SAM2~\cite{sam} into masks. Final benchmark triplets (image, instruction, mask) are human-validated.}
\label{fig:pipeline}
\end{figure}

With the rapid advancement of image editing models, existing benchmarks are increasingly insufficient to characterize editing performance in complex scenarios involving multiple similar instances. To address this limitation, we introduce a benchmark for measuring Multi-Instance Regional Alignment, \bench, designed for compositional instruction-based editing in multi-instance settings. 
Since our goal is to create a challenging benchmark with images containing multiple similar instances, there are limited real images that fulfill this criteria. Hence, we adopt an automatic dataset creation pipeline with human oversight and validation as illustrated in Fig.~\ref{fig:pipeline}. We generate both the reference images and the editing instructions, while ensuring high diversity and quality for the selected benchmark samples.

\noindent\textbf{Reference images.}
We construct a comprehensive prompt for generating a list of image descriptions depicting multiple instances of the same object. We find that using a multi-stage prompt composition to generate the descriptions leads to greater diversity. The prompt pipeline incorporates the following constraints: 1) the images contain three, four, or five instances of the same object; 2) the instances are clearly referable by a textual expression; 3) each image contains a unique combination of scenes, objects (including humans, vehicles, animals, common objects), and an otherwise rich environment, preventing bias towards frequently occurring categories. To adopt a progressively increasing difficulty, 50\% of the dataset contains three similar instances in each image, while those with four and five similar instances each account for 25\%. The full prompt is shown in the Supplementary (Fig.~\ref{fig:prompts_benchmark_image_description}).

To further improve cross-sample diversity, we automatically deduplicate and filter similar descriptions. Specifically, each generated description is compared against previously accepted ones using an LLM, and discarded if deemed semantically similar. For every filtered description, a new one is regenerated. Details of this filtering process are provided in Sec.~\ref{sec:supp_desc_gen} of the Supplementary.
We use Qwen3-14B-Instruct~\cite{qwen3-llm} to generate 200 image descriptions and we then use FLUX.2 [Dev]~\cite{flux.2} to synthesize reference images from these descriptions, with examples shown in Fig.~\ref{fig:example_descriptions}.
\begin{figure}[t]
\centering
\includegraphics[width=\columnwidth]{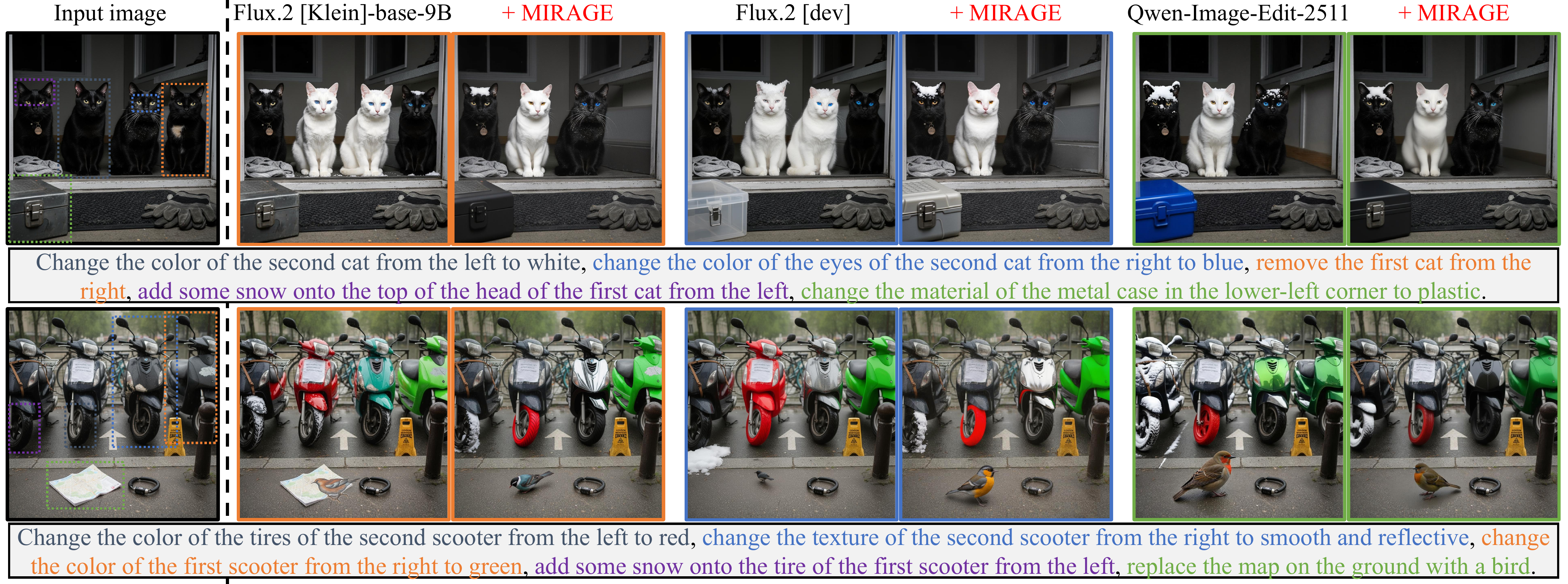}
\caption{Example images and instructions from \bench. The multiple similar instances and compositional instructions pose a challenge to SOTA models which introduce unintended modification in these complex scenarios. Through \method, we achieve precise instance-level editing while preserving background consistency.}
\label{fig:examples}
\end{figure}

\noindent\textbf{Edit instructions.}
We construct a prompt to generate editing instructions with Qwen3-VL-8B~\cite{qwen3-vl} based on both the reference image and the source description.
For each image, we require exactly five editing instructions and enforce instance-level one-to-one correspondence. Concretely, Qwen3-VL first orders the $n$ similar instances according to their horizontal spatial positions in the image, and assigns the first $n$ instructions to the corresponding instances from left to right. Any remaining instructions are restricted to other clearly visible and previously unedited objects in the scene. The edit instructions can always be unambiguously grounded and span five core types: addition, removal, replacement, color modification, and material modification. The full prompt for creating the instructions is provided in the Supplementary (Fig.~\ref{fig:prompts_benchmark_instruction}).

\noindent\textbf{Object masks.}
Together with the edit instructions, we let Qwen3-VL-8B extract the referring expression of the relevant object of each instruction. Using these referring expressions, we subsequently employ Qwen3-VL-8B to predict the corresponding bounding boxes of each object. These bounding boxes are then provided to SAM2~\cite{sam} to obtain precise target segmentation masks, which will be used during evaluation.

Following this pipeline, we generate 200 image–instruction samples and further conduct a human validation step to sub-select and retain a ``gold set'' of 100 high-quality samples that satisfy both structural consistency and semantic correctness. Fig.~\ref{fig:examples} illustrates example images and instructions from our benchmark (additional examples are shown in Fig.~\ref{fig:example_mybench}).
During evaluation, editing models must perform all five editing actions simultaneously through a single prompt. The resulting \bench systematically evaluates existing image editing models under multi-instance independent editing scenarios.

\section{\methodlong}
\label{sec:method}

\begin{figure}[t]
  \centering
  \includegraphics[width=0.9\columnwidth]{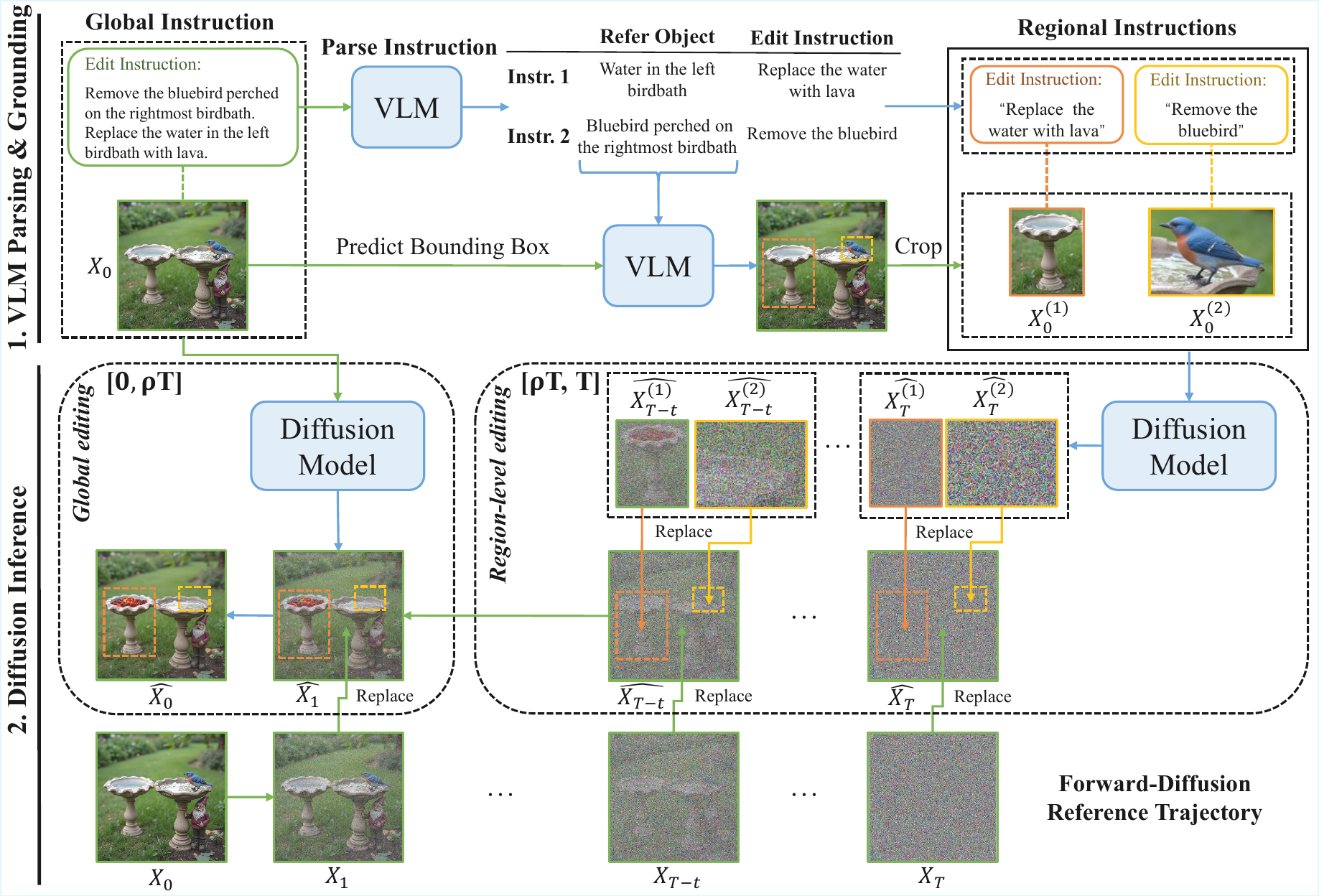}
  \caption{\textbf{Overview of \method.} (1) A VLM parses the global instruction into instance-level edits and localizes target regions via bounding boxes. (2) Regional branches perform parallel diffusion editing and inject their latents into the global branch; later time steps continue global editing under the global instruction while preserving background latents from the reference trajectory.}
  
  \label{fig:overview}
\end{figure}

While prior methods improve the image editing ability of base diffusion models through fine-tuning~\cite{train_flux,train_flux2,train_flux3}, we design a training-free framework to address the over-editing~\cite{wys,foi} and spatial misalignment~\cite{referringedit,pathiraja2025refedit} of current instruction-guided editing models, especially in complex scenarios. Our proposed framework, \methodlong (\method), is illustrated in Fig.~\ref{fig:overview}, and consists of two core components: 1) parsing the editing instruction to localize the corresponding target regions (Sec.~\ref{sec:method_loc}), and 2) designing an inference strategy tailored to the target and background regions (Sec.~\ref{sec:method_edit}).

\subsection{Instruction Parsing and Target Region Localization}
\label{sec:method_loc}

We leverage the capabilities of a visual-language model (VLM), such as Qwen3-VL, to perform instruction decomposition and target grounding without any additional model training.
As illustrated in Fig.~\ref{fig:overview} (top), given an input image \(x\) and its compositional editing instruction \(I\), we first employ the VLM as an information-extraction parser \(\mathcal{D}\) to decompose \(I\) into a set of aligned subtasks:
\begin{equation}
\mathcal{D}(I)=\{(r_k, I^k)\}_{k=1}^{K},
\end{equation}
where \(r_k\) denotes the referring expression of the \(k\)-th editing target, and \(I^k\) represents the corresponding atomic editing instruction.

Subsequently, we employ the VLM as a referring expression localizer \(\mathcal{L}\), which grounds the target specified by \(r_k\) in image \(x\) and extracts the corresponding sub-region based on the predicted bounding box:
\begin{gather}
\mathbf{b}_k = \mathcal{L}(x, r_k) \label{eq:bbox} \\
x^k = \mathrm{Crop}(x, \mathbf{b}_k) \label{eq:crop}
\end{gather}
where \(x^k\) denotes the region of interest extracted by cropping (Crop) the image $x$ with the predicted bounding box \(\mathbf{b}_k\).
This process produces a set of region-level editing instances
$\{(x^{k}, I^{k})\}_{k=1}^{K}$, which explicitly align editing regions with their intended operations, providing structured inputs for our multi-branch diffusion-based editing. The prompts used for both the instruction decomposition and the referring expression grounding are detailed in the Supplementary (Fig.~\ref{fig:prompts_locate_refer}).

\subsection{Localized image editing}
\label{sec:method_edit}

This module constitutes the editing component of our framework, implemented as a multi-branch parallel inference, built upon instruction-based editing models, such as FLUX.2~\cite{flux.2} and Qwen-Image-Edit~\cite{qwen-edit}. As illustrated in Fig.~\ref{fig:overview} (bottom), given the \(K\) region-level editing pairs \(\{(x^{k}, I^{k})\}_{k=1}^{K}\) obtained from the VLM, we utilize one global branch together with \(K\) target-region branches for parallel denoising.
Prior studies have shown that early diffusion time steps primarily determine editing semantics, whereas later time steps focus on refining local details~\cite{p2p,earlytimestep,null_text}. Motivated by this observation, we adopt a two-stage region fusion strategy during inference. Let the total number of time steps be \(T\), with $t=T$ corresponding to noise and $t=0$ the clean image, and denote the threshold time step for switching between region editing and global editing as \(\rho T\). 

Since we have isolated the edit regions of interest in the image, we can construct the noised latents of the background region from the forward diffusion process applied to the reference image, obtaining \({z}^{{ref}}_{t}\).
During the early stage (\(t > \rho T\)), the global branch is not used,
while each region editing branch $k$ independently performs denoising conditioned on \((x^{k}, I^{k})\), producing local latent representations \(z^{k}_{t}\). These local latents are then spatially aligned to their locations on the full image latents via the spatial mapping \(\Pi(\cdot, \mathbf{b}_k)\) and overwrite the respective locations of the background representation \({z}^{{ref}}_{t}\) to form the updated latents at time step \(t\):
\begin{equation}
z^{f}_{t}
=
{z}^{{ref}}_{t}
\odot
\left(1-\sum_{k=1}^{K} M_k\right)
+
\sum_{k=1}^{K}
\left(
\Pi(z^{k}_{t}, \mathbf{b}_k)
\odot M_k
\right)
\quad t > \rho T
\end{equation}
Here, \(M_k\) represents the binary mask, corresponding to bounding box \(\mathbf{b}_k\), projected onto the latent grid, and $\odot$ is the element-wise product. The fused latent \({z}^{{f}}_{t}\) is then propagated to the next denoising step.

During the late stage (\(t \le \rho T\)), all region branches are terminated, and only the global branch performs conditional denoising in the global latent space conditioned on \((x, I)\), producing \(\bar{z}^{f}_{t}\). In this stage, we continue to apply updates only within the target regions, while background representations are directly computed from the forward process of the reference image:
\begin{equation}
z^{f}_{t}
=
{z}^{{ref}}_{t}
\odot
\left(1-\sum_{k=1}^{K} M_k\right)
+
\bar{z}^{f}_{t}
\odot
\sum_{k=1}^{K} M_k
\quad t \le \rho T
\end{equation}
Here, $\bar z^{f}_{t}$ denotes the denoised latent predicted by the global branch before region-wise fusion.

In summary, during the early stage, the denoised latent representations from each region editing branch are written back to the corresponding locations of the reference latents, explicitly binding sub-instructions to their associated regions and enabling early semantic fusion. Throughout the entire denoising process, regions outside the targets consistently follow the reference trajectory, preventing unintended background drift. This inference strategy effectively mitigates over-editing, detail inconsistency, and spatial misalignment commonly observed in existing editing models~\cite{flux.1,flux.2,qwen-edit} under multi-instance and compositional instruction scenarios. Moreover, since regional edits are performed on a subset of the whole image, our \method pipeline is more efficient than editing with the base model as the diffusion model is applied on fewer patch tokens.

\section{Experiments}

We evaluate the effectiveness of \method in addressing the challenges of multi-instance and compositional image editing. We provide a quantitative (Sec.~\ref{sec:exp_quantitative}) and qualitative (Sec.~\ref{sec:exp_qualitative}) comparison against SOTA models on \bench and RefEdit-Bench. Finally, we conduct ablation studies (Sec.~\ref{sec:exp_ablation}) to analyze the impact $\rho$, our replacement strategies, the instruction complexity, and runtime.

\subsection{Experimental Setup}

\textbf{Datasets.} We conduct systematic comparative evaluations on the RefEdit-Bench~\cite{pathiraja2025refedit} single-instruction benchmark (200 images) and our proposed multi-instance and compositional instruction benchmark, \bench (100 images). Both datasets provide explicit referring expressions for each editing target, enabling precise localization and fine-grained evaluation.

\noindent\textbf{Baselines.} We compare against representative classical and SOTA instruction-based image editing models, including FLUX.2 [Dev/Klein]~\cite{flux.2}, Qwen-Image-Edit-2511~\cite{qwen-edit}, RefEdit(SD3)~\cite{pathiraja2025refedit}, MagicBrush~\cite{magicbrush}, and also GPT-Image-1.5~\cite{gpt-image} to systematically assess the effectiveness of our method.

\noindent\textbf{Metrics.} We partition each image into target and background regions using ground-truth masks provided by the benchmarks, and evaluate performance from two aspects: the degree to which the background region is preserved relative to the original image, and the extent to which the target region follows the editing instruction. 
We adopt EditScore~\cite{editscore} which follows the evaluation framework of VIEScore~\cite{ku2024viescore}, and provides a fine-tuned scoring model specifically for evaluating image editing, approximating human judgment.
EditScore evaluates results along three dimensions (on a scale from 0 - 10): Prompt Following (PF) measures whether the edit complies with the given instruction. Consistency (Cons) assesses whether unedited regions remain unchanged. Perceptual Quality (PQ) reflects overall image naturalness and the absence of artifacts. PF and Cons utilize the object masks from the benchmarks. The overall score is computed as $\mathrm{Overall}=\sqrt{\min(\mathrm{PF_{}},\mathrm{Cons_{}})\times\mathrm{PQ_{}}}$~\cite{editscore}.
To enhance evaluation reliability, we introduce finer-grained scoring criteria based on the original EditScore~\cite{editscore} evaluation template (see Fig.~\ref{fig:prompt_sc} and Fig.~\ref{fig:prompt_pq} for details).
PF and Cons are assessed using the Qwen3-VL-8B-Instruct variant of EditScore. However, as noted by Luo et al.~\cite{editscore}, the PQ score is less reliable which is why we resort to the stronger GPT-5.1 model to compute PQ.

\subsection{Quantitative Results}
\label{sec:exp_quantitative}

\begin{table}[t] %
\centering
\caption{\textbf{Results on \bench.} EditScore~\cite{editscore} for Prompt Following (PF) and Consistency (Cons), and GPT-5.1 evaluation of Perceptual Quality (PQ) in the range from 0 (lowest) to 10 (highest).}
\setlength{\tabcolsep}{12pt}
\label{tab:mybench_editscore}
\begin{adjustbox}{max width=\columnwidth,center}
\begin{tabular}{l c c c c}
\toprule
\textbf{Model}
& \makecell{\textbf{PF}$_{}\uparrow$\\ avg@3}
& \makecell{\textbf{Cons}$_{}\uparrow$\\ avg@3}
& \makecell{\textbf{PQ}$_{\mathrm{avg}}\uparrow$}
& \makecell{\textbf{Overall}$_{\mathrm{avg}}\uparrow$} \\
\midrule
MagicBrush~\cite{magicbrush}  & 0.879 & 9.788 & 7.648 & 2.358 \\
RefEdit-SD3~\cite{pathiraja2025refedit} & 2.358 & 8.025 & 6.324 & 4.247 \\
\hdashline
FLUX.2 [Klein]-9B~\cite{flux.2} & 6.046 & 8.646 & \textbf{8.988} & 7.372 \\
\rowcolor{lightblue} FLUX.2 [Klein]-9B + \textbf{MIRAGE} & \textbf{7.709} & \textbf{8.796} & 8.836 & \textbf{8.253} \\
\hdashline
FLUX.2 [Dev]~\cite{flux.2} & 6.982 & 8.378 & \textbf{8.880} & 7.874 \\
\rowcolor{lightblue} FLUX.2 [Dev] + \textbf{MIRAGE} & \textbf{8.086} & \textbf{9.006} & 8.808 & \textbf{8.439} \\
\hdashline
Qwen-Image-Edit-2511~\cite{qwen-edit} & 7.541 & 8.492 & \textbf{8.632} & 8.068 \\
\rowcolor{lightblue} Qwen-Image-Edit-2511 + \textbf{MIRAGE} & \textbf{8.004} & \textbf{8.850} & 8.436 & \textbf{8.217} \\
\hdashline
GPT-Image-1.5~\cite{gpt-image} & 7.659 & 8.894 & 9.000 & 8.302 \\
\bottomrule
\end{tabular}
\end{adjustbox}
\end{table}

\begin{table}[t] %
\centering
\caption{\textbf{Results on RefEdit-Bench.} EditScore~\cite{editscore} for Prompt Following (PF) and Consistency (Cons), and GPT-5.1 evaluation of Perceptual Quality (PQ) in the range from 0 (lowest) to 10 (highest).}
\setlength{\tabcolsep}{12pt}
\label{tab:quantitative_refedit_editscore}
\begin{adjustbox}{max width=\columnwidth,center}
\begin{tabular}{l c c c c}
\toprule
\textbf{Model}
& \makecell{\textbf{PF}$_{}\uparrow$\\ avg@3}
& \makecell{\textbf{Cons}$_{}\uparrow$\\ avg@3}
& \textbf{PQ}$_{\mathrm{avg}}\uparrow$
& \textbf{Overall}$_{\mathrm{avg}}\uparrow$ \\
\midrule
MagicBrush~\cite{magicbrush}  & 2.400 & 9.674 & 7.400 & 6.412 \\
RefEdit-SD3~\cite{pathiraja2025refedit} & 5.556 & 8.460 & 7.506 & 4.244 \\
\hdashline
FLUX.2 [Klein]-base-9B~\cite{flux.2} & 7.862 & 9.290 & \textbf{9.198} & 8.504 \\
\rowcolor{lightblue} FLUX.2 [Klein]-base-9B + \textbf{MIRAGE} & \textbf{8.236} & \textbf{9.296} & 9.066 & \textbf{8.641} \\
\hdashline
FLUX.2 [Dev]~\cite{flux.2} & 7.454 & 8.934 & \textbf{9.054} & 8.215 \\
\rowcolor{lightblue} FLUX.2 [Dev] + \textbf{MIRAGE} & \textbf{8.212} & \textbf{9.469} & 8.936 & \textbf{8.566} \\
\hdashline
Qwen-Image-Edit-2511~\cite{qwen-edit} & 8.208 & 9.182 & \textbf{8.928} & 8.560 \\
\rowcolor{lightblue} Qwen-Image-Edit-2511 + \textbf{MIRAGE} & \textbf{8.753} & \textbf{9.353} & 8.700 & \textbf{8.726} \\
\bottomrule
\end{tabular}
\end{adjustbox}
\end{table}

In Table~\ref{tab:mybench_editscore}, we evaluate baselines and \method on \bench. 
We perform three independent evaluations of PF and Cons for every image-instruction pair and calculate their average scores. The results indicate that, after integrating \method, all image editing models achieve significant improvements in instruction following (up to +1.7) and consistency (up to +0.7). This shows that \method enables precise editing of the correct target object and does not over-edit the image at incorrect regions. We observe that perceptual quality remains largely the same. We could not find a notable degradation in image quality in FLUX.2 models when using \method compared to the base models during qualitative inspection of the results, attributing the difference in PQ to model variance.

When applying \method to Qwen-Image-Edit-2511~\cite{qwen-edit}, we observe that the reduced PQ is explained by some edit results exhibiting slight blurring and loss of fine details in the target region, which becomes more noticeable for ``remove'' instructions. This is because Qwen-Image-Edit-2511~\cite{qwen-edit} resamples input images as a preprocessing step to a fixed resolution.
Since \method performs editing on locally cropped regions during the early denoising steps, this forced resampling can sometimes cause texture smoothing and loss of detail when small regions are upscaled significantly, thereby affecting perceptual quality. Hence, \method works best with editing models that allow variable size input images without rescaling, such as the FLUX.2~\cite{flux.2} model family.
Furthermore, we evaluated GPT-Image-1.5~\cite{gpt-image}, a proprietary SOTA model which performs significantly better than any base open-source model. In complex multi-instance editing scenarios, we find that GPT-Image-1.5~\cite{gpt-image} can still suffer from spatial misalignment and over-editing. Notably, \method applied to FLUX.2 [Dev] achieves a higher PF (8.09 vs. 7.66) and Cons (9.01 vs. 8.89) scores than GPT-Image-1.5 (Overall: 8.44 vs. 8.30), while FLUX.2 [Klein-9B] and Qwen-Image-Edit-2511 achieve performance very close to it.

Similarly, the results on RefEdit-Bench in Table~\ref{tab:quantitative_refedit_editscore} show that \method generalizes to other image editing benchmarks. While recent image editing models already perform well in these simpler scenarios with single instructions and fewer similar instances, \method further improves PF and Cons consistently across models.
In addition, we show in the Supplementary (Tab.~\ref{tab:supp_traditional_metric} and Tab.~\ref{tab:supp_quantitative_refedit_bg}) 
that integrating \method consistently improves traditional metrics on background preservation and CLIP image similarity across all backbone models, indicating that our approach effectively retains unedited regions of the reference image.

\subsection{Qualitative Results} %
\label{sec:exp_qualitative}

\begin{figure}[t]
\centering
\includegraphics[width=\columnwidth]{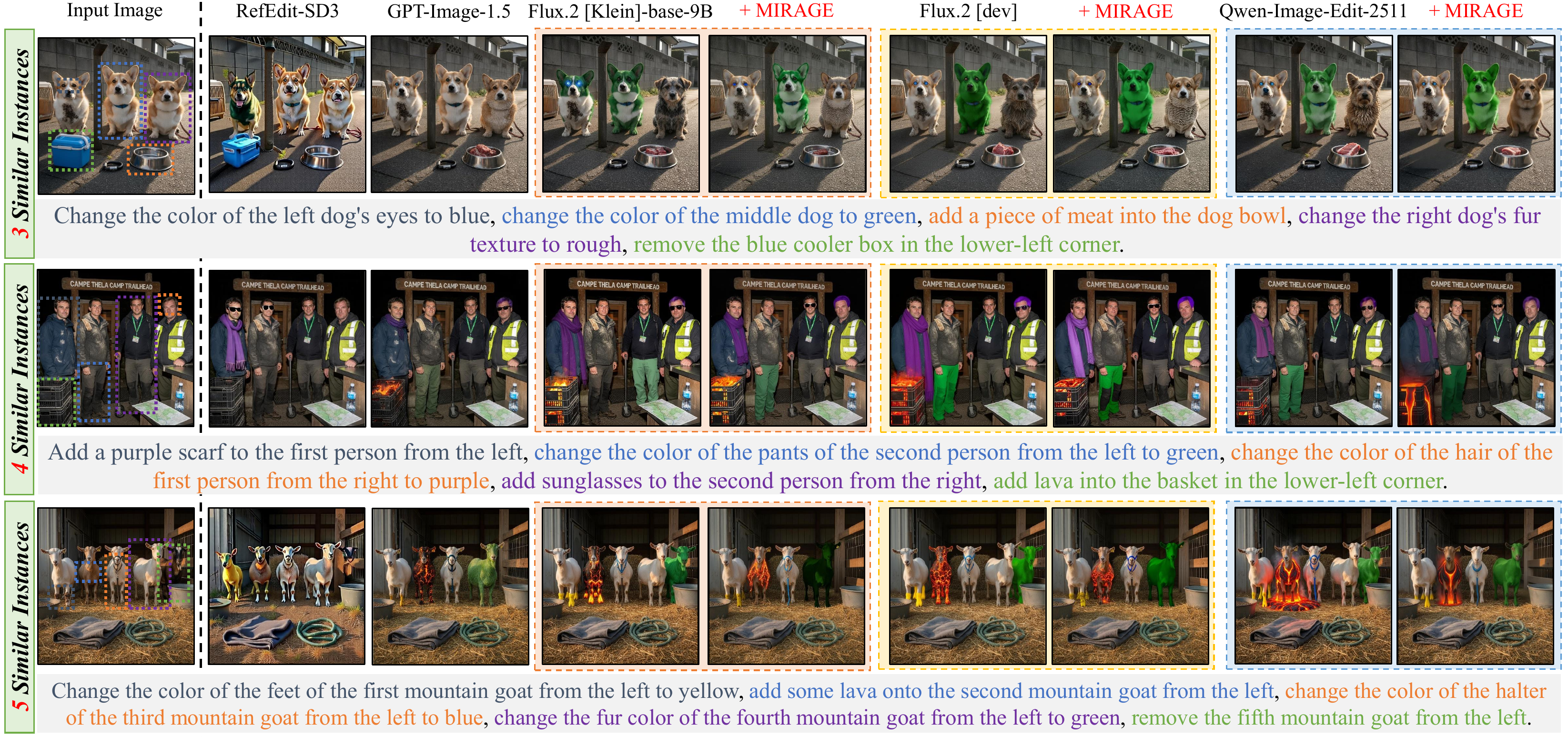}
\caption{\textbf{Qualitative results on \bench.} Integrating \method into SOTA models effectively mitigates over-editing in multi-instance scenarios while preserving detail consistency as much as possible.
}
\label{fig:mybench_results}
\end{figure}

Fig.~\ref{fig:mybench_results} presents qualitative results of baselines and \method on \bench. In complex scenarios involving multiple similar instances and compositional instructions, RefEdit-SD3 fails to produce correct edits. Although recent SOTA models generally follow the instructions well, they frequently exhibit severe over-editing and spatial misalignment, such as unintentionally changing the color of the left dog (first row, FLUX.2 [Klein]) or assigning sunglasses to the wrong person (second row, all models except with \method).

After integrating \method, the models can accurately identify the target instance specified by the instruction and perform precise, localized edits while preserving background consistency. For instance, no base model other than when combined with \method correctly removes the fifth goat in the last row. Moreover, qualitative results on RefEdit-Bench (Fig.~\ref{fig:supp_refedit_results}) 
further show that integrating \method effectively suppresses over-editing in the original models and consistently improves editing quality.

\subsection{Ablation Study}
\label{sec:exp_ablation}

\begin{figure}[t]
\centering
\begin{minipage}{0.55\textwidth}
  \centering
  \includegraphics[width=\linewidth]{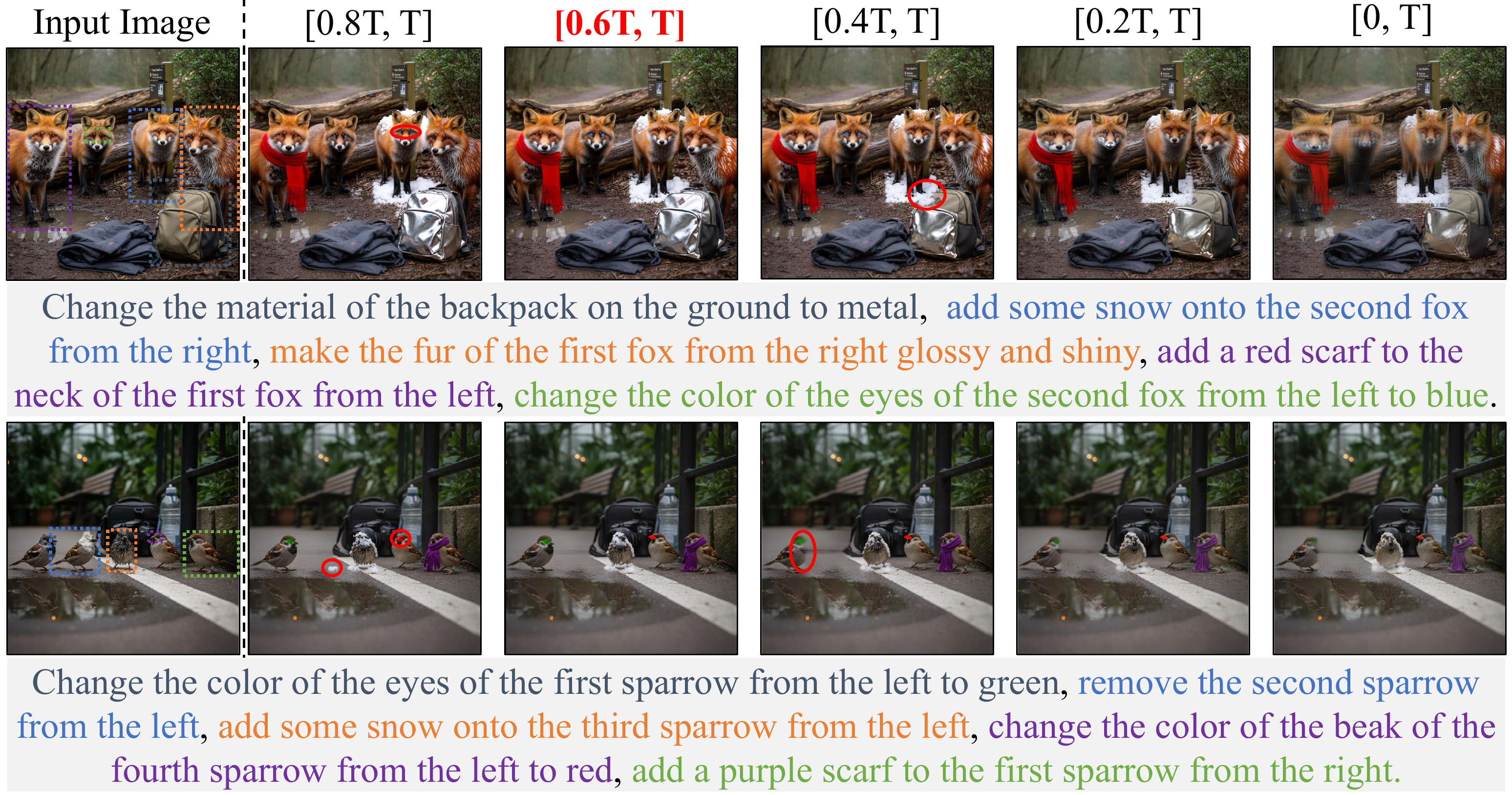}
\end{minipage}
\hfill
\begin{minipage}{0.42\textwidth}
  \centering
  \setlength{\tabcolsep}{4pt}
  \renewcommand{\arraystretch}{1.1}

  \label{tab:timestep}

    \resizebox{\linewidth}{!}{%
    \begin{tabular}{lcccc}
    \toprule
    \textbf{Time Step}
    & \makecell{\textbf{PF}$_{}\uparrow$\\ avg@3}
    & \makecell{\textbf{Cons}$_{}\uparrow$\\ avg@3}
    & \makecell{\textbf{PQ}$_{\mathrm{avg}}\uparrow$ }
    & \makecell{\textbf{Overall}$_{\mathrm{avg}}\uparrow$} \\
    \midrule
    $[0.8\mathrm{T},\,\mathrm{T}]$ & 7.917 & 8.945 & \textbf{8.848} & 8.370 \\
    \textcolor{red}{$[0.6\mathrm{T},\,\mathrm{T}]$} & \textbf{8.086} & \textbf{9.006} & 8.808 & \textbf{8.439} \\
    $[0.4\mathrm{T},\,\mathrm{T}]$ & 7.946 & 8.969 & 8.276 & 8.109 \\
    $[0.2\mathrm{T},\,\mathrm{T}]$ & 7.971 & 9.004 & 7.820 & 7.895 \\
    $[\mathrm{0},\,\mathrm{T}]$    & 7.899 & 8.936 & 6.368 & 7.092 \\
    \bottomrule
    \end{tabular}%
    }
\end{minipage}
\caption{\textbf{Effect of latent replacement time step $\rho$ on \bench}. Results obtained with FLUX.2 [Dev]. Red circles highlight artifacts or over-editing regions.
}
\label{fig:timestep}
\end{figure}

\textbf{Impact of $\rho T$ Time Step Choice.} We ablate the $\rho$ hyperparameter which determines when we switch from region editing to global editing. The quantitative results in Fig.~\ref{fig:timestep} on \bench indicate that our method is robust to the exact choice of $\rho$ in terms of PF and Cons metrics.
However, we observe that replacing target-region latents only within a short interval (e.g., [0.8T, T]) can sometimes be insufficient to effectively suppress over-editing. Expanding the replacement time-step range further alleviates over-editing and spatial misalignment, but at the cost of increased visual distortions and artifacts. As a result, we find that choosing the region editing interval to [0.6T, T] achieves the best trade-off between editing accuracy and visual quality. The analysis on RefEdit-Bench is provided in the Supplementary (Fig.~\ref{fig:ablation_timestep_refedit}).

\begin{figure}[t]
\centering
\begin{minipage}{0.55\columnwidth}
  \centering
  \includegraphics[width=\linewidth]{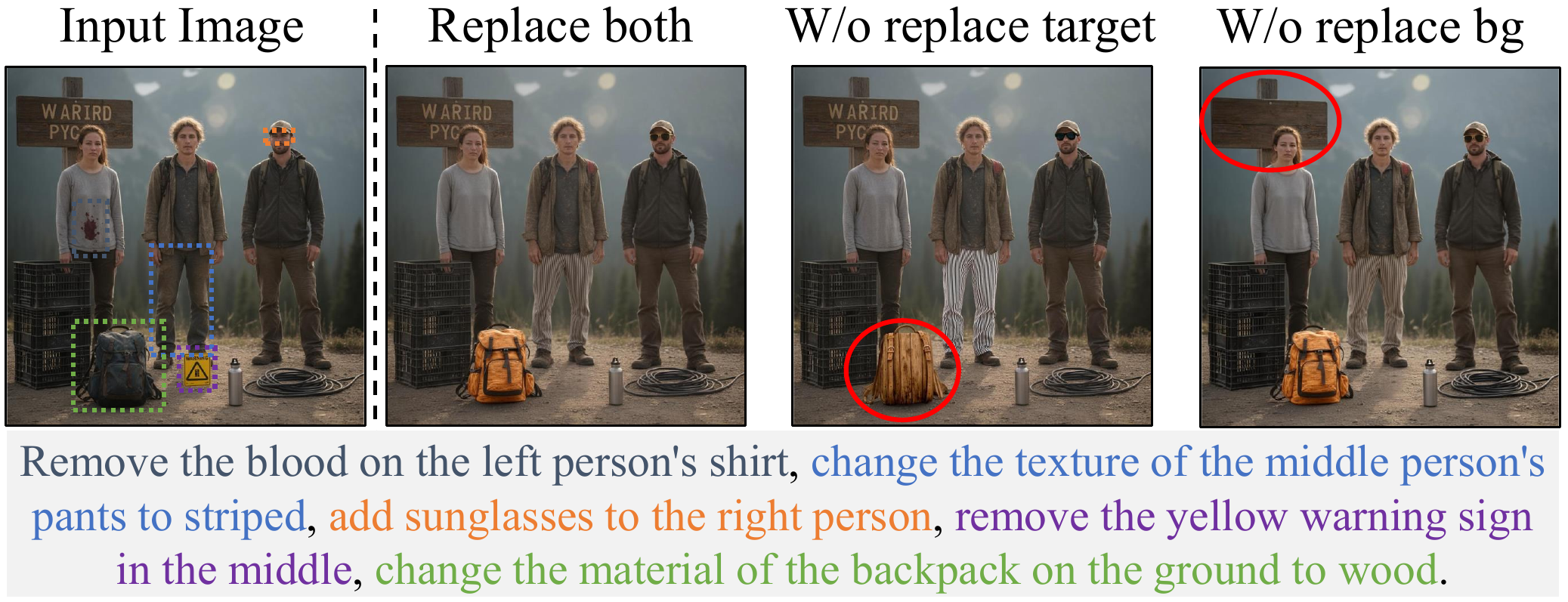}
\end{minipage}
\hfill
\begin{minipage}{0.42\columnwidth}
  \centering
  \setlength{\tabcolsep}{4pt}
  \renewcommand{\arraystretch}{1.1}
  \label{tab:strategy_ablation}
    \resizebox{\linewidth}{!}{
    \begin{tabular}{lcccc}
    \toprule
    \textbf{Strategy}
    & \makecell{\textbf{PF}$_{}\uparrow$\\ avg@3}
    & \makecell{\textbf{Cons}$_{}\uparrow$\\ avg@3}
    & \makecell{\textbf{PQ}$_{\mathrm{avg}}\uparrow$}
    & \makecell{\textbf{Overall}$_{\mathrm{avg}}\uparrow$}\\
    \midrule
    W/o replace target & 7.172 & 8.567 & 8.752 & 7.923 \\
    W/o replace bg     & 7.858 & 8.950 & \textbf{8.824} & 8.327 \\
    \textcolor{red}{Replace both}
    & \textbf{8.086} & \textbf{9.006} & 8.808 & \textbf{8.439} \\
    \bottomrule
    \end{tabular}
    }
\end{minipage}
\caption{\textbf{Effect of target and background replacement strategies on \bench.} These results are obtained with FLUX.2 [Dev] using a replacement interval of [0.6T, T]. Red circles indicate over-editing and local detail inconsistencies.}
\label{fig:strategy}
\end{figure}

\noindent\textbf{Impact of Target and Background Replacement Strategies.}
As described in Sec.~\ref{sec:method_edit}, our method consists of two core latent replacement strategies, namely the target-region replacement $\mathbf{R}_{\mathrm{target}}(t)= \sum_{k=1}^{K}\left(\Pi_k({z}^{k}_{t}, b_k)\odot {M}_k\right)$ and the background replacement $\mathbf{R}_{\mathrm{bg}}(t)= {z}^{{ref}}_{t}\odot\left(1-\sum_{k=1}^{K}M_k\right)$. To isolate their effects, we compare three configurations:
(1) \textbf{Replace both}, where both target and background replacements are enabled as defined above.
(2) \textbf{No target replacement}, where the multi-branch target latents are not injected during the early stage \(t>\rho T\). In this case, the target regions use the pre-fusion global prediction \(\bar{z}^{f}_{t}\), while the background replacement using \(z^{\mathrm{ref}}_{t}\) remains unchanged.
(3) \textbf{No background replacement}, where the background regions do not use the forward-diffusion reference latent \(z^{\mathrm{ref}}_{t}\) and instead use the pre-fusion global prediction \(\bar{z}^{f}_{t}\). The target-region replacement is preserved during the early stage \(t>\rho T\).
We show experimental results of this ablation in Fig.~\ref{fig:strategy}. Removing background replacement leads to structural drift in non-edited regions, as illustrated by the disappearance of the wooden sign text behind the woman in (fourth column). In contrast, removing target replacement weakens instruction following and results in reduced local detail consistency within edited regions (third column). The quantitative results on the right confirm these observations as the ``replace both'' strategy achieves the highest overall results.
Additional results on RefEdit-Bench (Fig.~\ref{fig:ablation_strategy_refedit}) 
corroborate that the two strategies exhibit complementary effects.

\begin{figure}[t]
\centering
\includegraphics[width=\columnwidth]{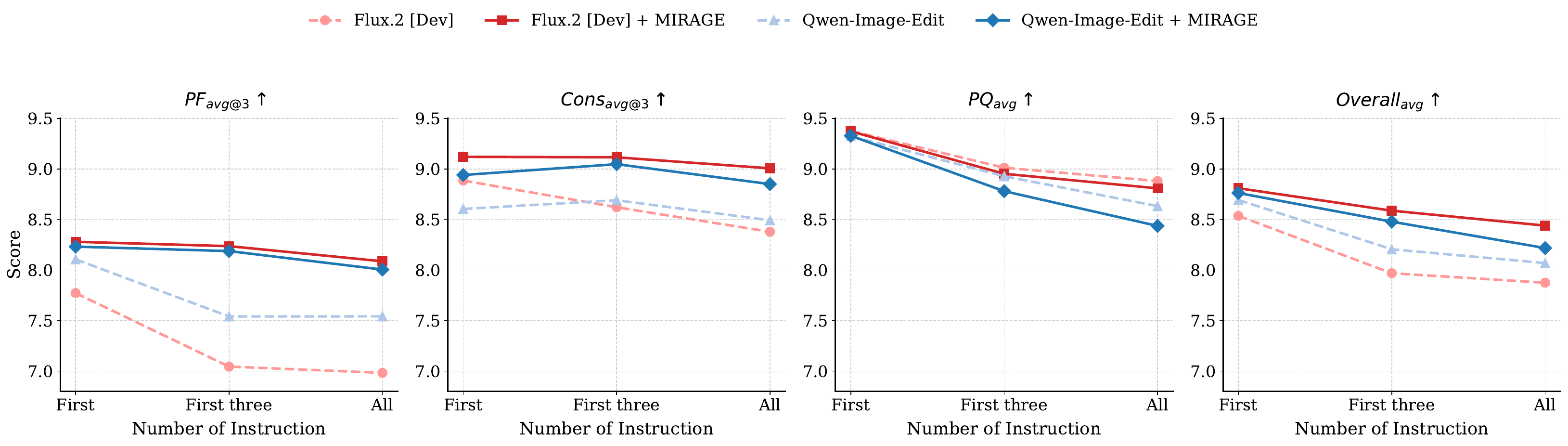}
\caption{\textbf{Performance under increasing instruction complexity.} Results on \bench using a replacement interval of $[0.6T,\,T]$.}
\label{fig:ins_num}
\end{figure}

\noindent\textbf{Instruction Complexity Analysis.}
We investigate the effect of increasingly more edit instructions on similar instances in \bench, essentially increasing the task difficulty. Fig.~\ref{fig:ins_num} shows the results when we vary the number of compositional instructions between one, three, and five.
While PF decreases for FLUX.2 [Dev] and Qwen-Image-Edit-2511 when going from one to three instructions, indicating that this poses an additional challenge to these models, it does not drop further with all five instructions. However, both models exhibit a steady decline in consistency (Cons) as the number of instructions increase, suggesting that accumulated over-editing becomes increasingly severe.
By comparison, after integrating \method, both FLUX.2 [Dev] and Qwen-Image-Edit-2511 demonstrate more stable performance trends across different complexity levels, particularly in terms of PF and Cons. These results further validate the effectiveness of \method in handling multi-instruction image editing scenarios.

\begin{table}[t]
\centering
\caption{\textbf{Runtime-performance comparison of different inference strategies on \bench.} Results obtained with FLUX.2 [Klein]-Base-9B under a replacement interval of [0.6T, T].} 
\setlength{\tabcolsep}{4pt}
\renewcommand{\arraystretch}{1.15}

\resizebox{\linewidth}{!}{
\begin{tabular}{lcccccccc}
\toprule
\textbf{Method} &
\makecell{\textbf{PF}$_{}\uparrow$\\ avg@3} &
\makecell{\textbf{Cons}$_{}\uparrow$\\ avg@3} &
\makecell{\textbf{PQ}$_{\mathrm{avg}}\uparrow$} &
\makecell{\textbf{Overall}$_{\mathrm{avg}}\uparrow$} &
\textbf{Parse (s)} &
\textbf{Detect (s)} &
\textbf{Inference (s)} &
\makecell{\textbf{Total}\\\textbf{Time}$\downarrow$} \\
\midrule

Flux.2 [Klein]-base-9B (Sequential) & 6.124 & 7.895 & 7.173 & 6.628 & -- & -- & 448.99 & 448.99 \\
Flux.2 [Klein]-base-9B              & 6.046 & 8.646 & \textbf{8.988} & 7.372 & -- & -- &  90.13 &  90.13 \\
Flux.2 [Klein]-base-9B + MIRAGE     & \textbf{7.709} & \textbf{8.796} & 8.836 & \textbf{8.253} & 2.90 & 8.37 & \textbf{74.59} & \textbf{85.87} \\

\bottomrule
\end{tabular}
}
\label{tab:runtime}

\end{table}

\noindent\textbf{Runtime and Performance Analysis.} We compare three inference strategies on \bench in terms of both efficiency and editing performance, including (i) standard execution using the base model, (ii) sequential execution that applies edit instructions one-by-one, and (iii) \method applied on top of the base model. All experiments are conducted on a single NVIDIA A100 40GB GPU.
As shown in Tab.~\ref{tab:runtime}, sequential execution requires approximately 5x longer inference time than standard base model execution, as expected. Although the sequential execution yields a slight improvement in PF, it leads to noticeable degradation in both Cons and PQ, due to compounding errors and high-frequency visual artifacts.
Qualitative examples are provided in Fig.~\ref{fig:sequential_experiments}
. In contrast, \method avoids the quality degradation while simultaneously improving all metrics at no additional computational cost. On the contrary, \method is overall slightly faster than base model execution because of the smaller latent size during the region editing phase. Instruction parsing and target localization introduce only about 11.3s overhead per image, while the main diffusion inference saves 15.5s.

\section{Conclusion}

We observe that current state-of-the-art models still struggle in complex scenarios involving multiple similar instances, where each instance requires independent editing. To address this limitation, we introduce \bench for assessing the performance of editing models on this task, and \method, 
a training-free framework designed to mitigate the over-editing and spatial misalignment issues.
Extensive experiments demonstrate that integrating \method into multiple backbone models consistently yields significant improvements in prompt following and consistency, validating the effectiveness and scalability of the proposed framework in complex editing scenarios. Additionally, \method is also computationally more efficient than using standard base model inference.

\noindent\textbf{Limitations.} The \method framework shares limitations with the VLM used to extract the referring object location and editing instructions. At the same time, we observe that even relatively small models, such as Qwen3-VL-8B as used by \method, perform well on these tasks. Moreover, the computational speedup of \method depends on the image editing model supporting variable input sizes for the images. FLUX.2 supports this, but Qwen Image Edit does not. Finally, since Qwen Image Edit always 
resizes the input image to a fixed size, we sometimes observe a slight degradation in image quality when applying \method due to blurring (Sec.~\ref{sec:supp_limitations} in Supplementary). This explains the slightly lower PQ scores when using Qwen Image Edit. However, FLUX.2 does not suffer from these limitations.

\subsubsection*{Acknowledgments.}
This work is supported by Hi! PARIS and ANR/France 2030 program (ANR-23-IACL-0005).

\bibliographystyle{splncs04}
\bibliography{main}

\begin{thebibliography}{10}
\providecommand{\url}[1]{\texttt{#1}}
\providecommand{\urlprefix}{URL }
\providecommand{\doi}[1]{https://doi.org/#1}

\bibitem{supp_pixtral2024}
Agrawal, P., et~al.: Pixtral 12b. arXiv preprint arXiv:2410.07073  (2024)

\bibitem{qwen3-vl}
Bai, S., Cai, Y., Chen, R., Chen, K., Chen, X., Cheng, Z., Deng, L., Ding, W., Gao, C., Ge, C., et~al.: Qwen3-vl technical report. arXiv preprint arXiv:2511.21631  (2025)

\bibitem{flux.1}
Batifol, S., Blattmann, A., Boesel, F., Consul, S., Diagne, C., Dockhorn, T., English, J., English, Z., Esser, P., et~al.: Flux. 1 kontext: Flow matching for in-context image generation and editing in latent space. arXiv preprint arXiv:2506.15742  (2025)

\bibitem{flux.2}
{Black Forest Labs}: {FLUX.2}: Analyzing and enhancing the latent space of {FLUX} -- representation comparison (2025), \url{https://bfl.ai/research/representation-comparison}

\bibitem{train_flux3}
Bradbury, R., Zhong, D.: Your latent mask is wrong: Pixel-equivalent latent compositing for diffusion models. arXiv preprint arXiv:2512.05198  (2025)

\bibitem{ip2p}
Brooks, T., Holynski, A., Efros, A.A.: Instructpix2pix: Learning to follow image editing instructions. In: CVPR (2023)

\bibitem{train_flux2}
Chen, J., Zhang, Y., Qian, X., Li, Z., Fermuller, C., Chen, C., Aloimonos, Y.: From inpainting to layer decomposition: Repurposing generative inpainting models for image layer decomposition. arXiv preprint arXiv:2511.20996  (2025)

\bibitem{DiffEdit}
Couairon, G., Verbeek, J., Schwenk, H., Cord, M.: Diffedit: Diffusion-based semantic image editing with mask guidance. In: ICLR (2023)

\bibitem{dit4edit}
Feng, K., Ma, Y., Wang, B., Qi, C., Chen, H., Chen, Q., Wang, Z.: Dit4edit: Diffusion transformer for image editing. In: AAAI (2025)

\bibitem{foi}
Guo, Q., Lin, T.: Focus on your instruction: Fine-grained and multi-instruction image editing by attention modulation. In: CVPR (2024)

\bibitem{p2p}
Hertz, A., Mokady, R., Tenenbaum, J., Aberman, K., Pritch, Y., Cohen-Or, D.: Prompt-to-prompt image editing with cross-attention control. In: ICLR (2023)

\bibitem{diffusion}
Ho, J., Jain, A., Abbeel, P.: Denoising diffusion probabilistic models. In: NeurIPS (2020)

\bibitem{supp_cogvlm2}
Hong, W., Wang, W., Ding, M., Yu, W., Lv, Q., Wang, Y., Cheng, Y., Huang, S., Ji, J., Xue, Z., et~al.: Cogvlm2: Visual language models for image and video understanding. arXiv preprint arXiv:2408.16500  (2024)

\bibitem{E4C}
Huang, T., Cao, P., Yang, L., Liu, C., Hu, M., Liu, Z., Song, Q.: E4c: Enhance editability for text-based image editing by harnessing efficient {CLIP} guidance. IEEE TCSVT  (2025)

\bibitem{earlytimestep}
Kim, J., Lee, Z., Cho, D., Jo, S., Jung, Y., Kim, K., Yang, E.: Early timestep zero-shot candidate selection for instruction-guided image editing. In: ICCV (2025)

\bibitem{FlexEdit}
Koo, G., Yoon, S., Hong, J.W., Yoo, C.D.: {FlexiEdit}: Frequency-aware latent refinement for enhanced non-rigid editing. In: ECCV (2024)

\bibitem{ku2024viescore}
Ku, M., Jiang, D., Wei, C., Yue, X., Chen, W.: Viescore: Towards explainable metrics for conditional image synthesis evaluation. In: ACL (2024)

\bibitem{zone}
Li, S., Zeng, B., Feng, Y., Gao, S., Liu, X., Liu, J., Li, L., Tang, X., Hu, Y., Liu, J., et~al.: Zone: Zero-shot instruction-guided local editing. In: CVPR (2024)

\bibitem{referringedit}
Liu, C., Li, X., Ding, H.: Referring image editing: Object-level image editing via referring expressions. In: CVPR (2024)

\bibitem{editscore}
Luo, X., Wang, J., Wu, C., Xiao, S., Jiang, X., Lian, D., Zhang, J., Liu, D., Liu, Z.: Editscore: Unlocking online {RL} for image editing via high-fidelity reward modeling. In: ICLR (2026)

\bibitem{wys}
Mirzaei, A., Aumentado-Armstrong, T., Brubaker, M.A., Kelly, J., Levinshtein, A., Derpanis, K.G., Gilitschenski, I.: Watch your steps: Local image and scene editing by text instructions. In: ECCV (2024)

\bibitem{supp_mistral2025large3}
{Mistral AI}: Mistral large 3 (2025), \url{https://mistral.ai/news/mistral-3}

\bibitem{null_text}
Mokady, R., Hertz, A., Aberman, K., Pritch, Y., Cohen-Or, D.: Null-text inversion for editing real images using guided diffusion models. In: CVPR (2023)

\bibitem{lazy-dit}
Nitzan, Y., Wu, Z., Zhang, R., Shechtman, E., Cohen-Or, D., Park, T., Gharbi, M.: Lazy diffusion transformer for interactive image editing. In: ECCV (2024)

\bibitem{gpt-image}
{OpenAI}: The new {ChatGPT} {Images} is here (2025), \url{https://openai.com/index/new-chatgpt-images-is-here/}

\bibitem{lov}
Patashnik, O., Garibi, D., Azuri, I., Averbuch-Elor, H., Cohen-Or, D.: Localizing object-level shape variations with text-to-image diffusion models. In: ICCV (2023)

\bibitem{stylegan}
Patashnik, O., Wu, Z., Shechtman, E., Cohen-Or, D., Lischinski, D.: {StyleCLIP}: Text-driven manipulation of {StyleGAN} imagery. In: ICCV (2021)

\bibitem{pathiraja2025refedit}
Pathiraja, B., Patel, M., Singh, S., Yang, Y., Baral, C.: Refedit: A benchmark and method for improving instruction-based image editing model on referring expressions. In: ICCV (2025)

\bibitem{dit}
Peebles, W., Xie, S.: Scalable diffusion models with transformers. In: ICCV (2023)

\bibitem{sdxl}
Podell, D., English, Z., Lacey, K., Blattmann, A., Dockhorn, T., M{\"u}ller, J., Penna, J., Rombach, R.: {SDXL}: Improving latent diffusion models for high-resolution image synthesis. In: ICLR (2024)

\bibitem{dalle2}
Ramesh, A., Dhariwal, P., Nichol, A., Chu, C., Chen, M.: Hierarchical text-conditional image generation with clip latents. arXiv preprint arXiv:2204.06125  (2022)

\bibitem{sam}
Ravi, N., Gabeur, V., Hu, Y.T., Hu, R., Ryali, C., Ma, T., Khedr, H., R{\"a}dle, R., Rolland, C., Gustafson, L., et~al.: {SAM} 2: Segment anything in images and videos. In: ICLR (2025)

\bibitem{groundingdino}
Ren, T., Jiang, Q., Liu, S., Zeng, Z., Liu, W., Gao, H., Huang, H., Ma, Z., Jiang, X., Chen, Y., et~al.: Grounding {DINO} 1.5: Advance the {Edge} of open-set object detection. arXiv preprint arXiv:2405.10300  (2024)

\bibitem{stablediffusion}
Rombach, R., Blattmann, A., Lorenz, D., Esser, P., Ommer, B.: High-resolution image synthesis with latent diffusion models. In: CVPR (2022)

\bibitem{gip2p}
Shagidanov, A., Poghosyan, H., Gong, X., Wang, Z., Navasardyan, S., Shi, H.: Grounded-instruct-pix2pix: Improving instruction based image editing with automatic target grounding. In: ICASSP (2024)

\bibitem{lime}
Simsar, E., Tonioni, A., Xian, Y., Hofmann, T., Tombari, F.: {LIME}: Localized image editing via attention regularization in diffusion models. In: WACV (2025)

\bibitem{supp_regionreasoner}
Sun, W., Chen, H., Du, Y., Zheng, Y., Snoek, C.G.M.: Regionreasoner: Region-grounded multi-round visual reasoning. In: ICLR (2026)

\bibitem{supp_unlocking}
Wang, J., Wu, Z., Huang, D., Zheng, Y., Wang, H.: Unlocking the potential of mllms in referring expression segmentation via a light-weight mask decoder. arXiv preprint arXiv:2508.04107  (2025)

\bibitem{instructedit}
Wang, Q., Zhang, B., Birsak, M., Wonka, P.: Instructedit: Improving automatic masks for diffusion-based image editing with user instructions. arXiv preprint arXiv:2305.18047  (2023)

\bibitem{supp_internvl3}
Wang, W., Gao, Z., Gu, L., Pu, H., Cui, L., Wei, X., Liu, Z., Jing, L., Ye, S., Shao, J., et~al.: Internvl3. 5: Advancing open-source multimodal models in versatility, reasoning, and efficiency. arXiv preprint arXiv:2508.18265  (2025)

\bibitem{omniEdit}
Wei, C., Xiong, Z., Ren, W., Du, X., Zhang, G., Chen, W.: Omniedit: Building image editing generalist models through specialist supervision. In: ICLR (2025)

\bibitem{qwen-edit}
Wu, C., Li, J., Zhou, J., Lin, J., Gao, K., Yan, K., Yin, S.m., Bai, S., Xu, X., Chen, Y., et~al.: Qwen-image technical report. arXiv preprint arXiv:2508.02324  (2025)

\bibitem{qwen3-llm}
Yang, A., Li, A., Yang, B., Zhang, B., Hui, B., Zheng, B., Yu, B., Gao, C., Huang, C., Lv, C., et~al.: Qwen3 technical report. arXiv preprint arXiv:2505.09388  (2025)

\bibitem{oir}
Yang, Z., Ding, G., Wang, W., Chen, H., Zhuang, B., Shen, C.: Object-aware inversion and reassembly for image editing. In: ICLR (2024)

\bibitem{imgEdit}
Ye, Y., He, X., Li, Z., Lin, B., Yuan, S., Yan, Z., Hou, B., Yuan, L.: Imgedit: A unified image editing dataset and benchmark. In: NeurIPS (2025)

\bibitem{magicbrush}
Zhang, K., Mo, L., Chen, W., Sun, H., Su, Y.: Magicbrush: A manually annotated dataset for instruction-guided image editing. In: NeurIPS (2023)

\bibitem{hive}
Zhang, S., Yang, X., Feng, Y., Qin, C., Chen, C.C., Yu, N., Chen, Z., Wang, H., Savarese, S., Ermon, S., et~al.: Hive: Harnessing human feedback for instructional visual editing. In: CVPR (2024)

\bibitem{train_flux}
Zhang, Z., Xie, J., Lu, Y., Yang, Z., Yang, Y.: Enabling instructional image editing with in-context generation in large scale diffusion transformer. In: NeurIPS (2025)

\end{thebibliography}

\appendix
\newpage

\setcounter{figure}{0}
\setcounter{table}{0}
\setcounter{equation}{0}

\renewcommand{\thefigure}{S\arabic{figure}}
\renewcommand{\thetable}{S\arabic{table}}
\renewcommand{\theequation}{S\arabic{equation}}

\begin{center}
{\fontsize{12pt}{14pt}\selectfont\bfseries MIRAGE: Benchmarking and Aligning\\Multi-Instance Image Editing\\-\\Supplementary Material\par}
\end{center}

\section{Implementation Details of \bench Construction}
\label{sec:supp_filter_process}

In this section we detail how we construct \bench, including the image description generation~\ref{sec:supp_desc_gen}, image synthesis~\ref{sec:supp_img_gen}, edit instruction and referring object generation~\ref{sec:supp_instr_gen}, and mask generation~\ref{sec:supp_mask_gen}.

\subsection{Image Description Generation}
\label{sec:supp_desc_gen}

We use Qwen3-14B-Instruct~\cite{qwen3-llm} for image description generation. However, prompting the model naively to generate image descriptions leads to template collapse, i.e., repetitive descriptions which often lack sufficient detail, fail to satisfy the referring requirements, or producing the same instance-scene combinations across samples.

To improve the quality of the generated data, we adopt a two-stage generation strategy. In the first stage, we use the top prompt shown in Fig.~\ref{fig:prompts_benchmark_image_description} to ask the LLM to generate a large set of (repeated instance category, scene) pairs. If duplicate pairs are detected across samples, they are discarded and resampled until 200 valid pairs are obtained. During repeated sampling, the temperature and top\_p parameters are gradually increased according to the number of failures in order to encourage diversity. The default sampling parameters are restored once a valid pair is produced. Candidate image descriptions are then generated based on these valid pairs.

In the second stage, we first use the middle prompt in Fig.~\ref{fig:prompts_benchmark_image_description} together with the (repeated instance category, scene) pairs to generate image descriptions. Then, using the bottom prompt in Fig.~\ref{fig:prompts_benchmark_image_description}, the LLM automatically verifies the generated descriptions. The checking process includes instance count, referential distinguishability, completeness of visual details, and duplication across samples. If a candidate does not satisfy these constraints, the model provides revision suggestions and performs up to four rounds of refinement. Samples that still fail after the refinement rounds are discarded and regenerated from the first stage. Examples of the final generated image descriptions can be found in the "Description" row of Fig.~\ref{fig:example_descriptions}.

\subsection{Image Synthesis}
\label{sec:supp_img_gen}

\begin{figure}[t] %
\centering
\includegraphics[width=\columnwidth]{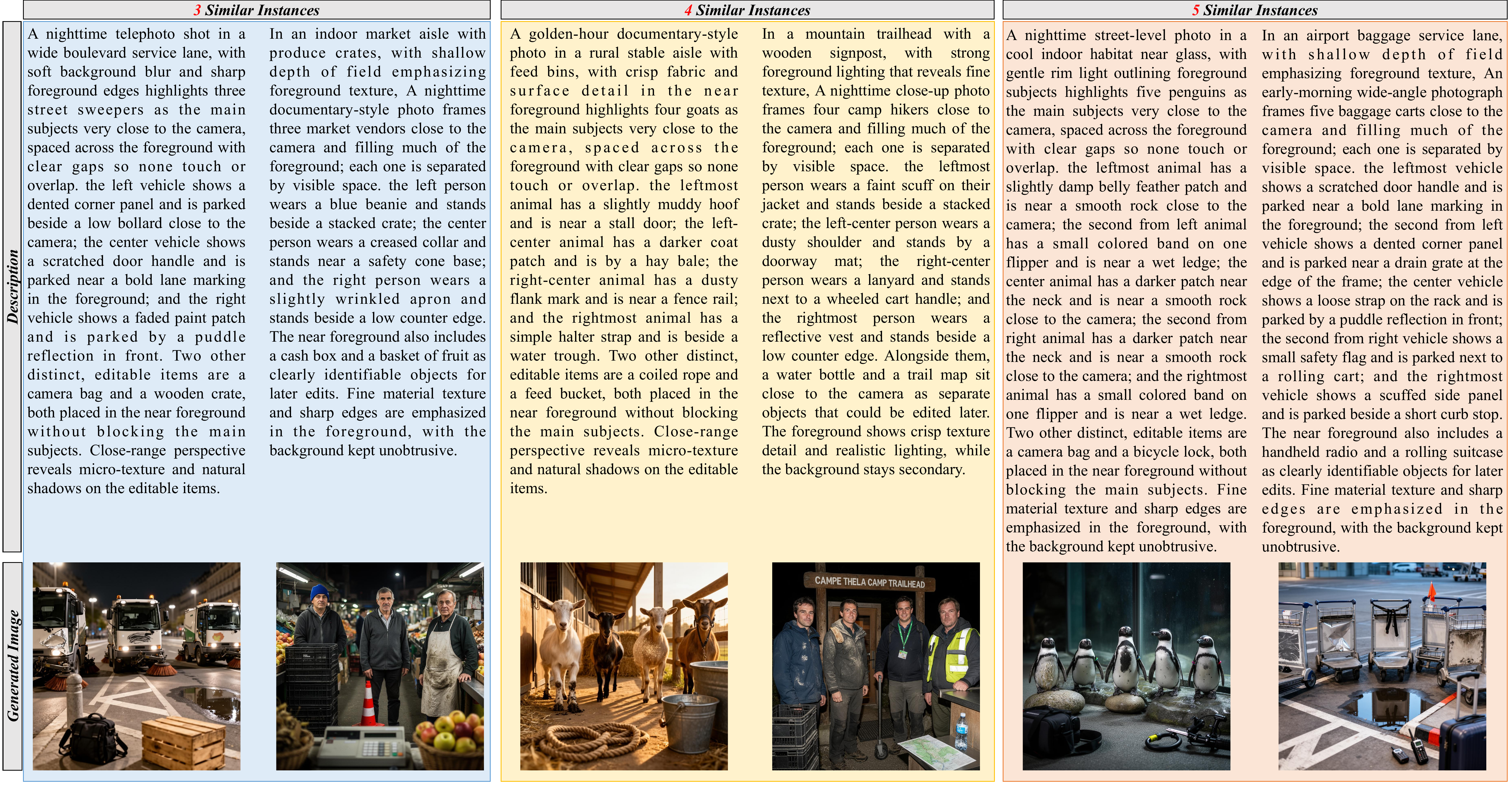}
\caption{\textbf{The Image descriptions and the corresponding generated image of \bench.}}
\label{fig:example_descriptions}
\end{figure}

The generated descriptions are used as prompts for the text-to-image model FLUX.2 [Dev]~\cite{flux.2} to synthesize images. The images generated from the image descriptions are shown in the "Generated Image" row of Fig.~\ref{fig:example_descriptions}. Most importantly, multi-instance scenes can sometimes lead to unstable generations, for example incorrect instance counts, objects extending outside the image boundary, or insufficient visual detail. Therefore we manually inspect the generated results and select 100 samples with clearer semantics and higher visual quality from the 200 candidates to form the final \bench dataset. 

\subsection{Editing Instruction and Referring Object Generation}
\label{sec:supp_instr_gen}

Here we use Qwen3-VL-8B-Instruct~\cite{qwen3-llm} for instruction and refer object generation. Compared to image description generation, editing instruction generation is less prone to template collapse, and therefore no additional judge stage is required. The generation of editing instructions and referring objects also follows a two-stage process.
In the first stage, we use the top prompt in Fig.~\ref{fig:prompts_benchmark_instruction} to construct a slot plan based solely on the filtered image descriptions. The slot plan records the number of repeated instances, the object category, and key attribute details. This information serves as a weak prior for later instruction generation.

In the second stage, using the middle prompt in Fig.~\ref{fig:prompts_benchmark_instruction}, the VLM generates editing instructions conditioned on both the slot plan and the synthesized image. The model is explicitly instructed to rely primarily on the image content and the final generated editing instructions are shown in the "Edit Instruction" row of Fig.~\ref{fig:example_mybench}. In addition, the final referring objects are obtained by parsing the generated editing instructions using the bottom prompt in Fig.~\ref{fig:prompts_benchmark_instruction}. 

\subsection{Ground-Truth Mask Generation}
\label{sec:supp_mask_gen}

After obtaining the referring objects in each image, we first use a Qwen3-VL-8B-Instruct~\cite{qwen3-vl} to localize the target regions and predict their corresponding bounding boxes. The predicted bounding boxes are then used as prompts for SAM2~\cite{sam} to generate the target segmentation masks. We note that models from the SAM family were tested in our pipeline, and based on visual inspection of the generated results, SAM2~\cite{sam} shows more reliable segmentation quality and stability. It is therefore used to produce the final masks. Each sample in \bench finally consists of three components: the original image, the editing instruction, and the corresponding GT mask. The final results of \bench are shown in Fig.~\ref{fig:example_mybench}.

\section{\bench Examples}
\label{sec:supp_mybench_generated_example}

Fig.~\ref{fig:example_descriptions} shows example image descriptions generated for \bench, along with the corresponding synthesized images. Each description explicitly specifies multiple similar instances and distinctive attributes to ensure referential distinguishability in later editing tasks. 
The generated images confirm that these descriptions can be faithfully rendered while preserving the intended instance layout and visual details.

\begin{figure}[t] %
\centering
\includegraphics[width=\columnwidth]{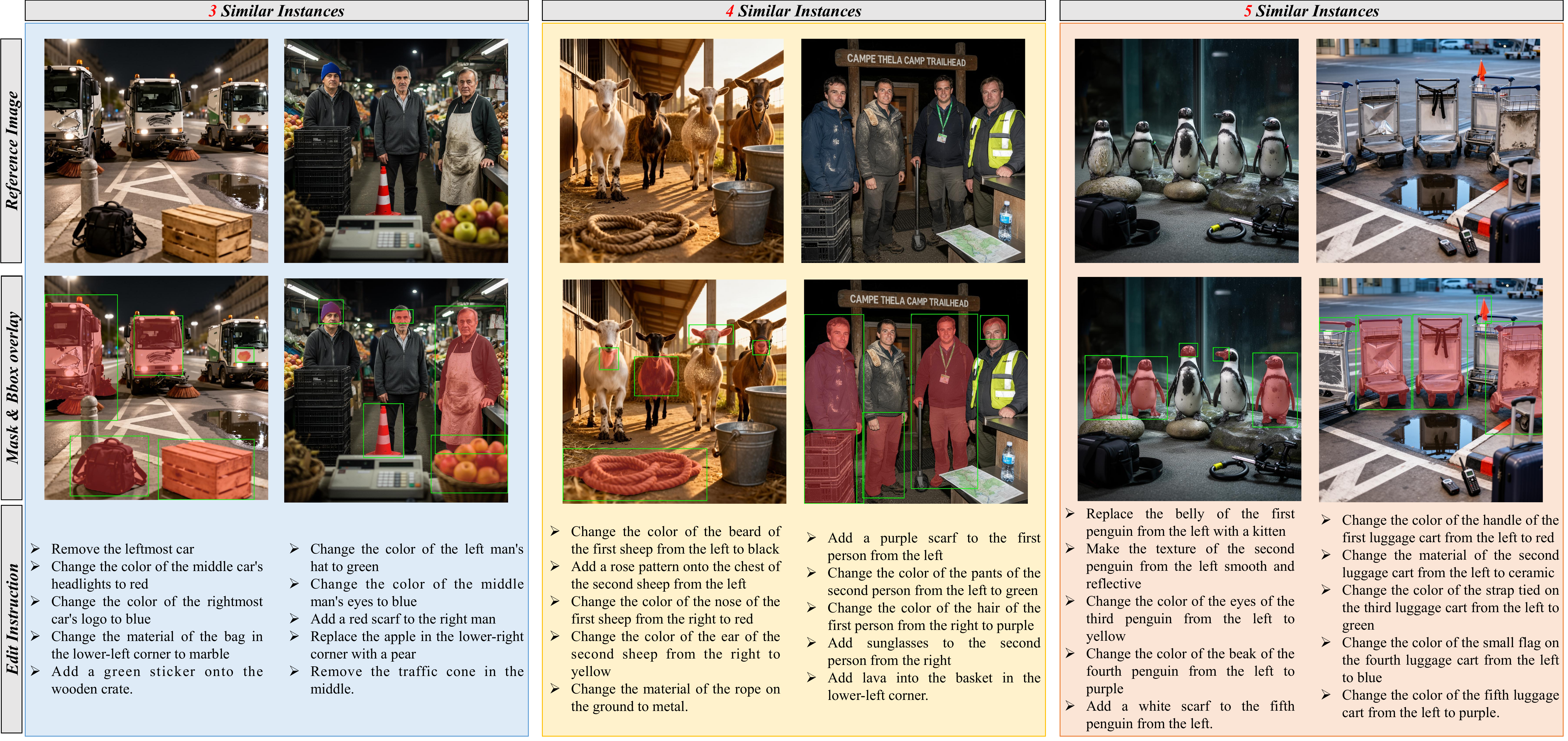}
\caption{\textbf{\bench sample examples.} The first row shows the synthesized original images, the second row presents the corresponding ground-truth (GT) masks of the target regions, and the third row displays the editing instructions constructed based on the generated image semantics and the source prompts.}
\label{fig:example_mybench}
\end{figure}

Fig.~\ref{fig:example_mybench} shows several example entries from \bench, where each column corresponds to one editing example constructed in the benchmark. The first row presents the synthesized source images, the second row shows the corresponding ground-truth (GT) masks that indicate the target editing regions, and the third row lists the editing instructions generated based on the image semantics and the source prompts.
These examples illustrate the multi-instance editing scenarios covered in \bench, where target objects are specified through referring expressions and localized accordingly. The ground-truth (GT) masks are further used for evaluating the editing performance in subsequent metrics.

\section{Additional Results}

\subsection{Additional Quantitative Metrics}
\label{sec:supp_all_table_results}

\begin{table}[t] %
\centering
\caption{\textbf{Additional evaluation metrics on \bench.}}
\label{tab:supp_traditional_metric}
\begin{adjustbox}{max width=\columnwidth,center}
\begin{tabular}{l c c c c c c c}
\toprule
\multirow{2}{*}{\textbf{Model}} 
& \multirow{2}{*}{\makecell{\textbf{Structure} \\ \textbf{Distance}$\downarrow$}}
& \multicolumn{4}{c}{\textbf{Background preservation}} 
& \multicolumn{2}{c}{\textbf{CLIP similarity}} \\
\cmidrule(lr){3-6} \cmidrule(lr){7-8}
& & \textbf{PSNR}$\uparrow$ & \textbf{LPIPS}$\downarrow$ 
& \textbf{MSE}$\downarrow$ & \textbf{SSIM}$\uparrow$ 
& \textbf{Whole}$\uparrow$ & \textbf{Edited}$\uparrow$ \\
\midrule
RefEdit-SD3 & 0.030 & 21.615 & 0.173 & 0.011 & 0.688 & 20.905 & 20.591 \\
MagicBrush  & 0.006 & 25.969 & 0.103 & 0.003 & 0.777 & 19.027 & 20.516 \\
\hdashline
FLUX.2 [Klein]-base-9B & \textbf{0.020} & 24.626 & 0.043 & 0.006 & 0.928 & 20.586 & 21.491 \\
\rowcolor{lightblue} FLUX.2 [Klein]-base-9B + \textbf{MIRAGE} & 0.027 & \textbf{25.514} & \textbf{0.038} & \textbf{0.003} & \textbf{0.912} & \textbf{21.137} & \textbf{21.958} \\
\hdashline
FLUX.2 [Dev] & 0.027 & 23.023 & 0.066 & 0.008 & 0.855 & 21.813 & 21.830 \\
\rowcolor{lightblue} FLUX.2 [Dev] + \textbf{MIRAGE} & 0.027 & \textbf{25.108} & \textbf{0.041} & \textbf{0.004} & \textbf{0.906} & \textbf{21.989} & \textbf{22.297} \\
\hdashline
Qwen-Image-Edit-2511 & 0.025 & 23.632 & 0.072 & 0.007 & 0.870 & 21.555 & 22.014 \\
\rowcolor{lightblue} Qwen-Image-Edit-2511 + \textbf{MIRAGE} & \textbf{0.019} & \textbf{28.729} & \textbf{0.040} & \textbf{0.002} &\textbf{ 0.918} & \textbf{21.569} & \textbf{22.160} \\
\bottomrule
\end{tabular}
\end{adjustbox}
\end{table}

\begin{table}[t] %
\centering
\caption{\textbf{Additional evaluation metrics on RefEdit-Bench}}
\label{tab:supp_quantitative_refedit_bg}

\begin{adjustbox}{max width=\columnwidth,center}
\begin{tabular}{l c c c c c c c}
\toprule
\multirow{2}{*}{\textbf{Model}} 
& \multirow{2}{*}{\makecell{\textbf{Structure} \\ \textbf{Distance}$\downarrow$}}
& \multicolumn{4}{c}{\textbf{Background preservation}} 
& \multicolumn{2}{c}{\textbf{CLIP similarity}} \\
\cmidrule(lr){3-6} \cmidrule(lr){7-8}
& & \textbf{PSNR}$\uparrow$ & \textbf{LPIPS}$\downarrow$ 
& \textbf{MSE}$\downarrow$ & \textbf{SSIM}$\uparrow$ 
& \textbf{Whole}$\uparrow$ & \textbf{Edited}$\uparrow$ \\
\midrule
RefEdit-SD3 & 0.024 & 24.463 & 0.072 & 0.010 & 0.867 & 23.108 & 21.229 \\
MagicBrush  & 0.015 & 25.287 & 0.069 & 0.009 & 0.784 & 21.412 & 20.350 \\
\hdashline
FLUX.2 [Klein]-base-9B & 0.023 & 26.927 & 0.044 & 0.007 & 0.924 & 22.063 & 21.349 \\
\rowcolor{lightblue} FLUX.2 [Klein]-base-9B + \textbf{MIRAGE} & \textbf{0.021} & \textbf{29.097} & \textbf{0.026} & \textbf{0.002} & \textbf{0.938} & \textbf{22.247} & \textbf{21.479} \\
\hdashline
FLUX.2 [Dev] & 0.032 & 25.349 & 0.075 & 0.012 & 0.888 & 22.342 & 21.302 \\
\rowcolor{lightblue} FLUX.2 [Dev] + \textbf{MIRAGE} & \textbf{0.021} & \textbf{28.742} & \textbf{0.028} & \textbf{0.002} & \textbf{0.935} & \textbf{22.698} & \textbf{22.110} \\
\hdashline
Qwen-Image-Edit-2511 & 0.022 & 25.331 & 0.066 & 0.006 & 0.811 & 22.424 & 21.686 \\
\rowcolor{lightblue} Qwen-Image-Edit-2511 + \textbf{MIRAGE} & \textbf{0.017} & \textbf{33.977} & \textbf{0.021} & \textbf{0.001} & \textbf{0.971} & \textbf{22.442} & \textbf{21.766} \\
\bottomrule
\end{tabular}
\end{adjustbox}
\end{table}

In addition to the EditScore results reported in the main paper~\cite{editscore}, we also evaluate several widely used image similarity metrics to provide a complementary analysis of editing quality, particularly for assessing background preservation. These metrics quantify different aspects of similarity between the edited image and the original source image. Specifically, PSNR and MSE measure pixel-level reconstruction differences, SSIM evaluates structural similarity, and LPIPS measures perceptual similarity in deep feature space. All metrics are computed on the background regions, defined as the areas outside the union of all ground-truth target masks, in order to evaluate how well non-target regions are preserved during editing.

From the background preservation metrics in Table~\ref{tab:supp_traditional_metric} and Table~\ref{tab:supp_quantitative_refedit_bg}, it can be observed that \method consistently improves PSNR and SSIM while reducing LPIPS and MSE across all backbone models. This indicates that \method effectively suppresses unnecessary modifications to non-target regions during diffusion-based editing, thereby enhancing background preservation and overall visual stability.
For example, on the relatively simpler RefEdit-Bench~\cite{pathiraja2025refedit}, although the base models already achieve reasonably strong performance, integrating \method still brings consistent improvements. Specifically, for FLUX.2 [Dev], the PSNR increases from 25.349 to 28.742, while the MSE decreases from 0.012 to 0.002. This further demonstrates that \method improves background consistency across different levels of task difficulty.

For the CLIP Edited metric, we apply the ground-truth target mask to the edited image to retain only the edited object region, and compute the image–text similarity between this local region and the corresponding fine-grained editing instruction. This metric evaluates the accuracy of the edit on the target object. And CLIP Whole measures the image–text similarity between the full edited image and the global editing instruction, assessing the overall semantic alignment of the edited result.
In addition, the CLIP similarity results show that after applying \method, the semantic similarity generally remains stable or slightly improves across models, indicating that the proposed method does not weaken the models’ ability to follow editing instructions while reducing unintended edits. 
It is worth noting that CLIP has known limitations in evaluating referring expressions~\cite{pathiraja2025refedit}. Therefore, this metric should only be considered as a complementary reference, while the primary evaluation results are reported in Tables~\ref{tab:mybench_editscore} and~\ref{tab:quantitative_refedit_editscore} in the main paper.

\subsection{Ablation on VLMs for Object Localization on \bench}
\begin{table}[t] %
\centering
\caption{\textbf{Ablation on different VLMs.}
We replace the VLM used for referring object localization in \method and results are reported on \bench}
\setlength{\tabcolsep}{12pt}
\label{tab:mybench_abalation_vlm}
\begin{adjustbox}{max width=\columnwidth,center}
\begin{tabular}{c c c c c c}
\toprule
\textbf{Model} & \textbf{Vision-Language Model}
& \makecell{\textbf{PF}$_{}\uparrow$\\ avg@3}
& \makecell{\textbf{Cons}$_{}\uparrow$\\ avg@3}
& \makecell{\textbf{PQ}$_{\mathrm{avg}}\uparrow$}
& \makecell{\textbf{Overall}$_{\mathrm{avg}}\uparrow$} \\
\midrule
\multirow{3}{*}{\makecell{FLUX.2 [Klein]-base-9B \\ + MIRAGE}}
& Qwen3-VL-8B-Instruct~\cite{qwen3-vl} & 7.709 & 8.796 & 8.836 & 8.253 \\
& Qwen3-VL-4B-Instruct~\cite{qwen3-vl} & 7.699 & 8.848 & 8.724 & 8.195 \\
& RegionReasoner~\cite{supp_regionreasoner} & 7.287 & 8.769 & 8.840 & 8.026 \\
\bottomrule
\end{tabular}
\end{adjustbox}
\end{table}

We conduct an ablation study by replacing the VLM used for referring object localization while keeping all other components of the pipeline unchanged. This experiment evaluates whether the effectiveness of \method depends on a specific VLM. All results are reported on \bench using the EditScore~\cite{editscore} metrics.

As shown in Table~\ref{tab:mybench_abalation_vlm}, Qwen3-VL 4B and 8B achieve similar PF scores, while RegionReasoner yields a lower PF. This is mainly due to its weaker referring localization capability, which leads to more frequent target-binding errors in complex scenes. Nevertheless, its performance still surpasses that of the basic model, indicating that \method consistently improves editing performance across different VLM choices.
Moreover, the Cons scores are very similar across different VLMs, suggesting that the region control mechanism in \method effectively suppresses spatial misalignment and over-editing among similar instances. Meanwhile, PQ remains largely stable across different VLMs, as perceptual quality is primarily determined by the underlying generative backbone. Overall, these results demonstrate that \method remains effective and robust under different VLM settings.

We additionally tested several other VLMs (e.g., InternVL3.5-8B~\cite{supp_internvl3}, CogVLM2\cite{supp_cogvlm2}, MLLMSeg\cite{supp_unlocking}, Mistral Large 3~\cite{supp_mistral2025large3}, and Pixtral Large~\cite{supp_pixtral2024}). In our experiments, these models in general underperform in producing stable referring localization outputs. 
As more VLMs with this capability become available, we expect that \method can be equally applied using them.

\subsection{Additional Qualitative Results on \bench}
\label{sec:supp_mybench_image_results}
\begin{figure}[t] 
\centering
\includegraphics[width=\columnwidth]{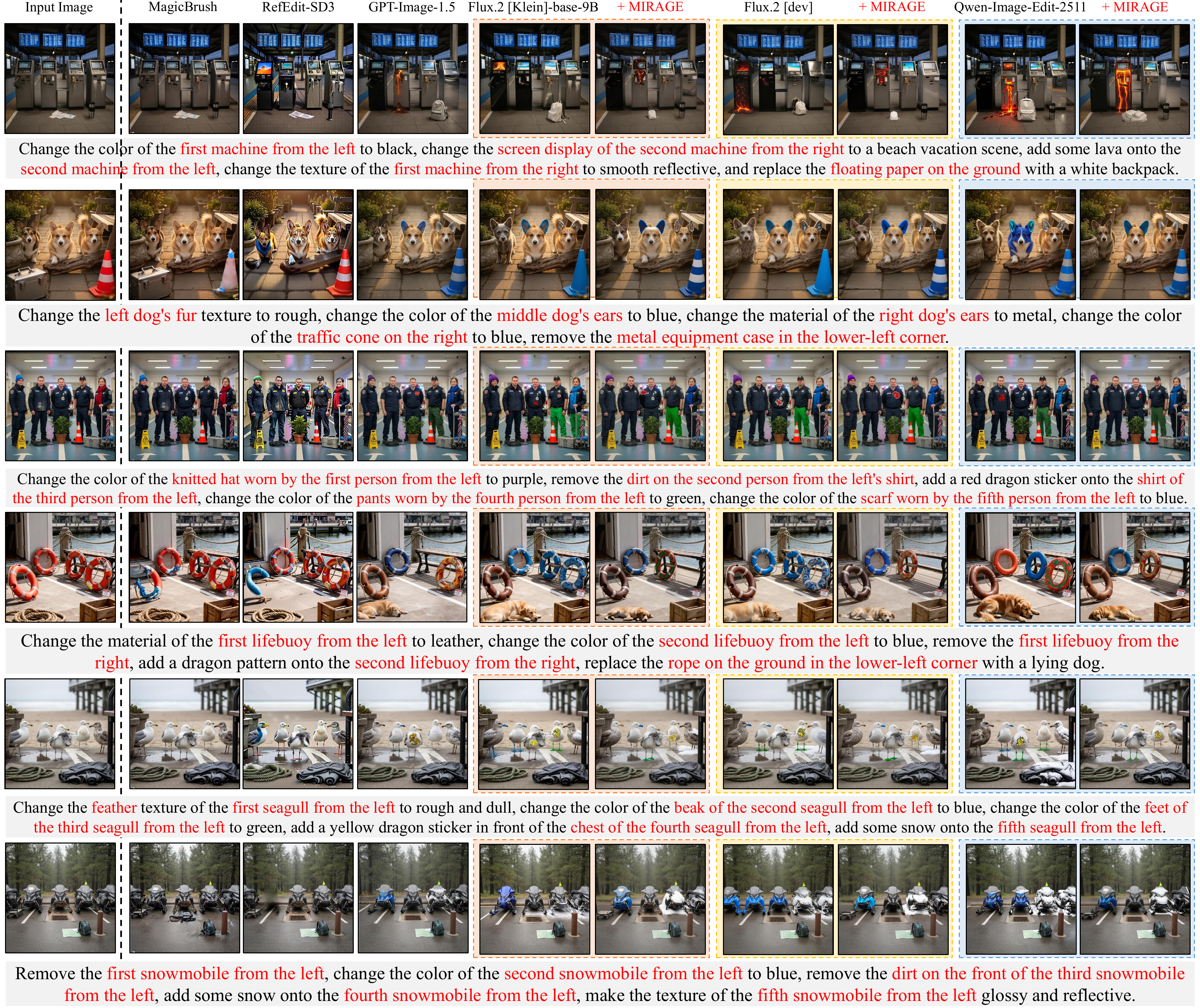}
\caption{\textbf{Additional qualitative results on \bench.}}
\label{fig:additional_results_mybench}
\end{figure}

\bench presents challenging scenarios with multiple similar instances and compositional instructions. As illustrated in Fig.~\ref{fig:additional_results_mybench}, classical editing models such as MagicBrush~\cite{magicbrush} and RefEdit-SD3~\cite{pathiraja2025refedit} often fail to follow the complicated instructions and produce severe distortions and structural degradation in this case.
More advanced models, such as GPT-5.1~\cite{gpt-image}, can execute most instructions and generate visually plausible results. However, they still exhibit common issues including spatial misalignment and over-editing. For example, in the fourth row the model incorrectly removes the third swim ring from the left and applies the dragon pattern to the wrong instance. In the fifth row, the model fails to recolor the feet of the third seagull from the left and instead places the dragon sticker on an incorrect instance. 

Overall, these examples demonstrate that accurately binding editing instructions to the correct target instance is highly challenging in multi-instance editing scenarios. By integrating \method into the backbone models, these issues can be effectively mitigated.

\subsection{Sequential vs. Joint Editing on \bench}
\label{Sec:supp_order}

\begin{figure}[t] %
\centering
\includegraphics[width=.8\columnwidth]{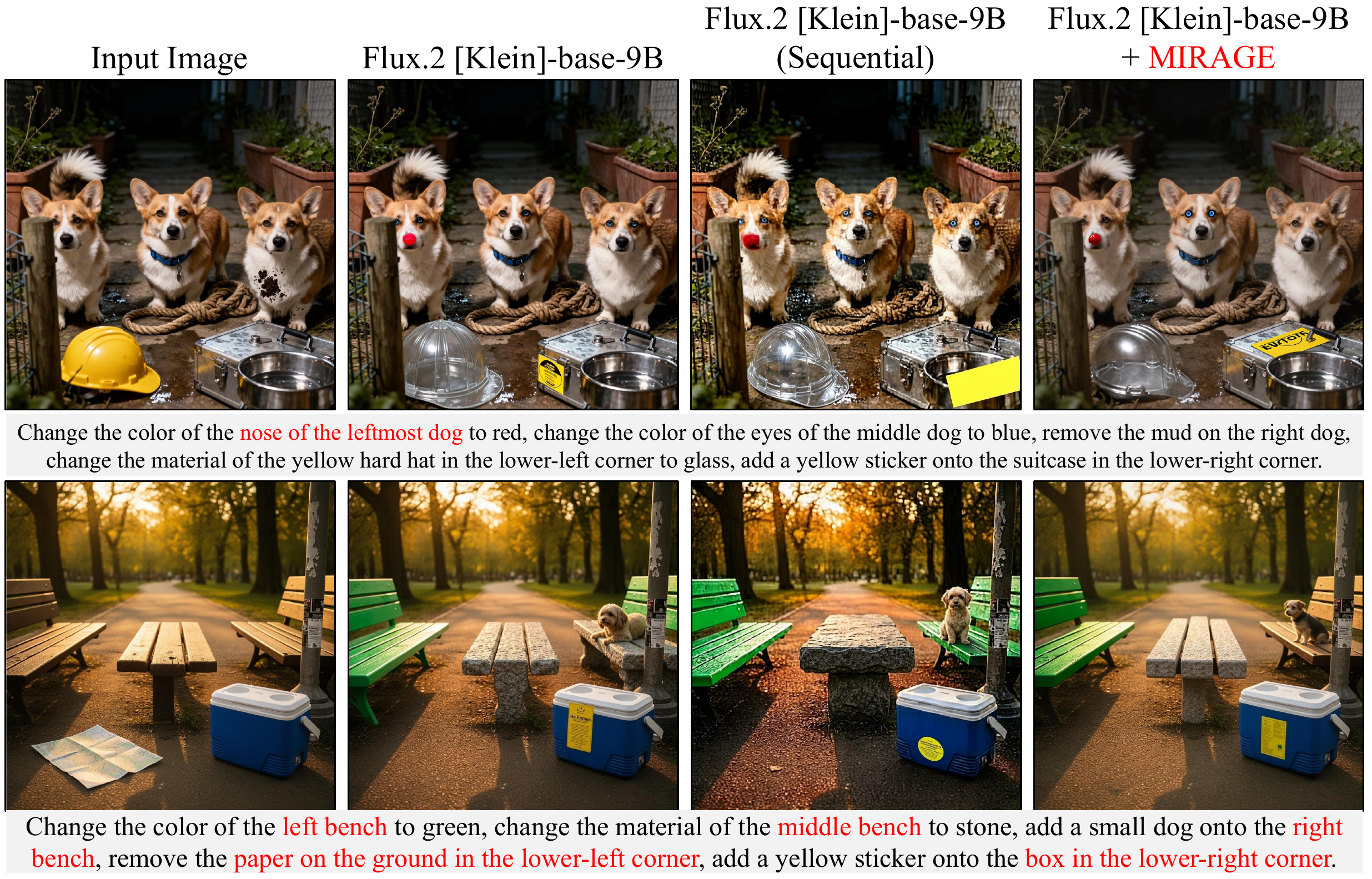}
\caption{\textbf{Sequential vs. joint editing inference.} Applying instructions sequentially leads to accumulated artifacts and structural inconsistencies due to repeated image re-editing. In contrast, MIRAGE performs joint multi-region inference within a single denoising trajectory, preventing error accumulation and maintaining global coherence.}
\label{fig:sequential_experiments}
\end{figure}

To investigate whether sequential editing is preferable for backbone models in multi-instance editing scenarios, we compare three strategies: sequential editing, joint editing with the backbone model, and joint editing with our method.

In sequential editing, instructions are applied one by one, with the edited image repeatedly fed back into the generative model as the input for the next step. However, this process tends to accumulate errors. 
As shown in Fig.~\ref{fig:sequential_experiments}, sequential editing accumulates errors, leading to oversharpening artifacts and structural distortions. More importantly, sequential execution on the basic model still produces similar over-editing as the joint editing baseline, indicating that sequential editing does not effectively mitigate this issue.
In contrast, \method performs joint multi-region editing within a single denoising trajectory. By resolving all target regions simultaneously, the model avoids repeated image re-editing and mitigates over-editing.

\subsection{Additional Qualitative Results on RefEdit-Bench}
\label{sec:supp_refedit_image_results}
\begin{figure}[t] %
\centering
\includegraphics[width=\columnwidth]{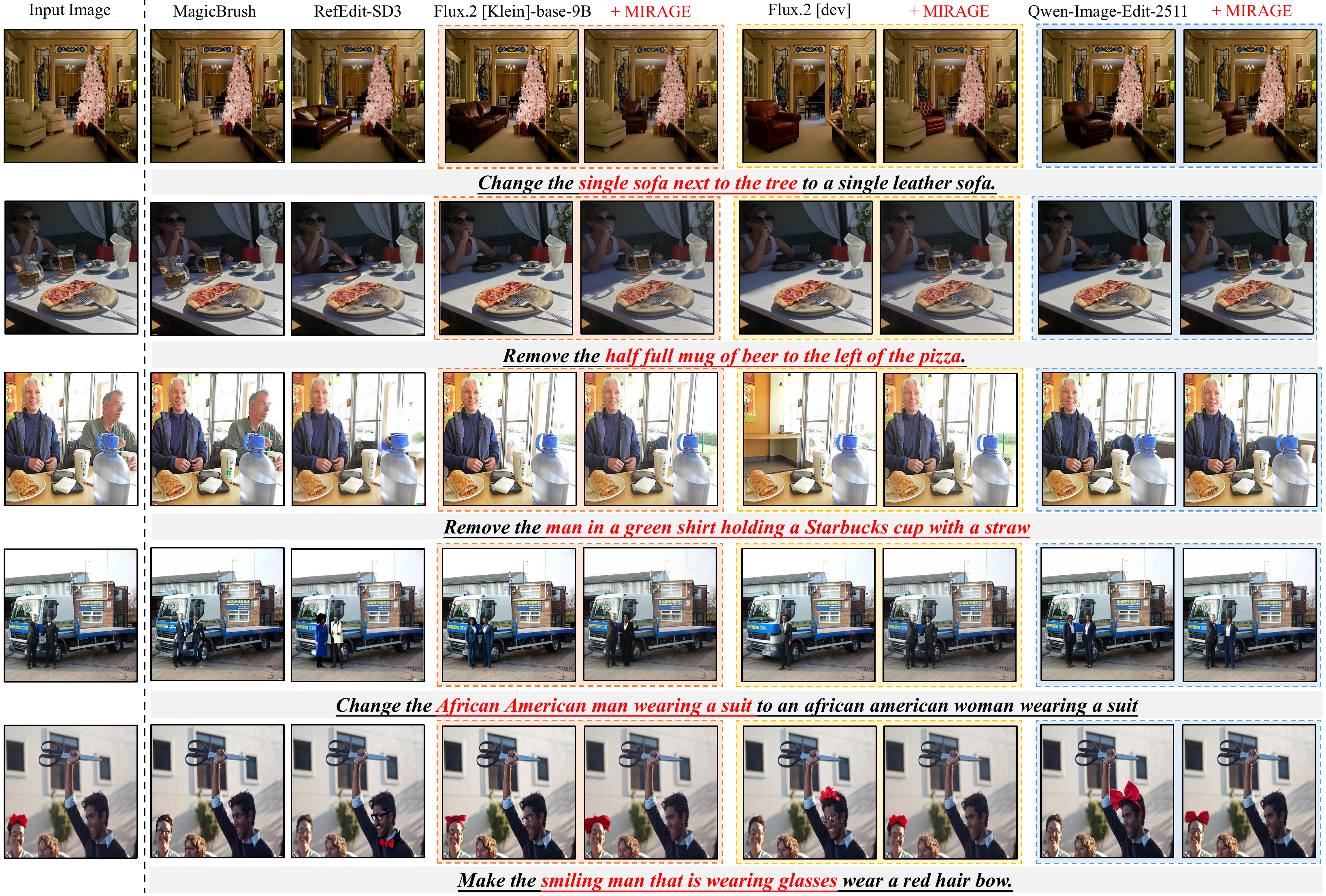}
\caption{\textbf{Qualitative results on RefEdit-Bench}. Across all compared methods, integrating our \method consistently improves performance, substantially outperforming both the baseline models and their original versions.}
\label{fig:supp_refedit_results}
\end{figure}

Fig.~\ref{fig:supp_refedit_results} presents qualitative comparisons on RefEdit-Bench~\cite{pathiraja2025refedit}. Among the baseline methods, MagicBrush~\cite{magicbrush} often struggles to correctly execute the editing instructions, while RefEdit-SD3~\cite{pathiraja2025refedit} performs relatively better but introduces visible artifacts (e.g., color noise in the lower-right example). For stronger editing models, such as FLUX.2~\cite{flux.2} and Qwen-Image-Edit~\cite{qwen-edit}, the instructions are generally followed, but noticeable over-editing is frequently observed in non-target regions.
After integrating \method, these issues are largely mitigated. The edited regions better match the intended instructions while the surrounding content remains more stable, resulting in improved visual consistency across different basic models.

\subsection{Ablation on Latent Replacement on RefEdit-Bench}
\subsubsection{Effect of Time-step Interval}
\label{Sec:ablation_timestep_refedit}

\begin{figure}[t] %
\centering
\begin{minipage}{0.55\columnwidth}
  \centering
  \includegraphics[width=\linewidth]{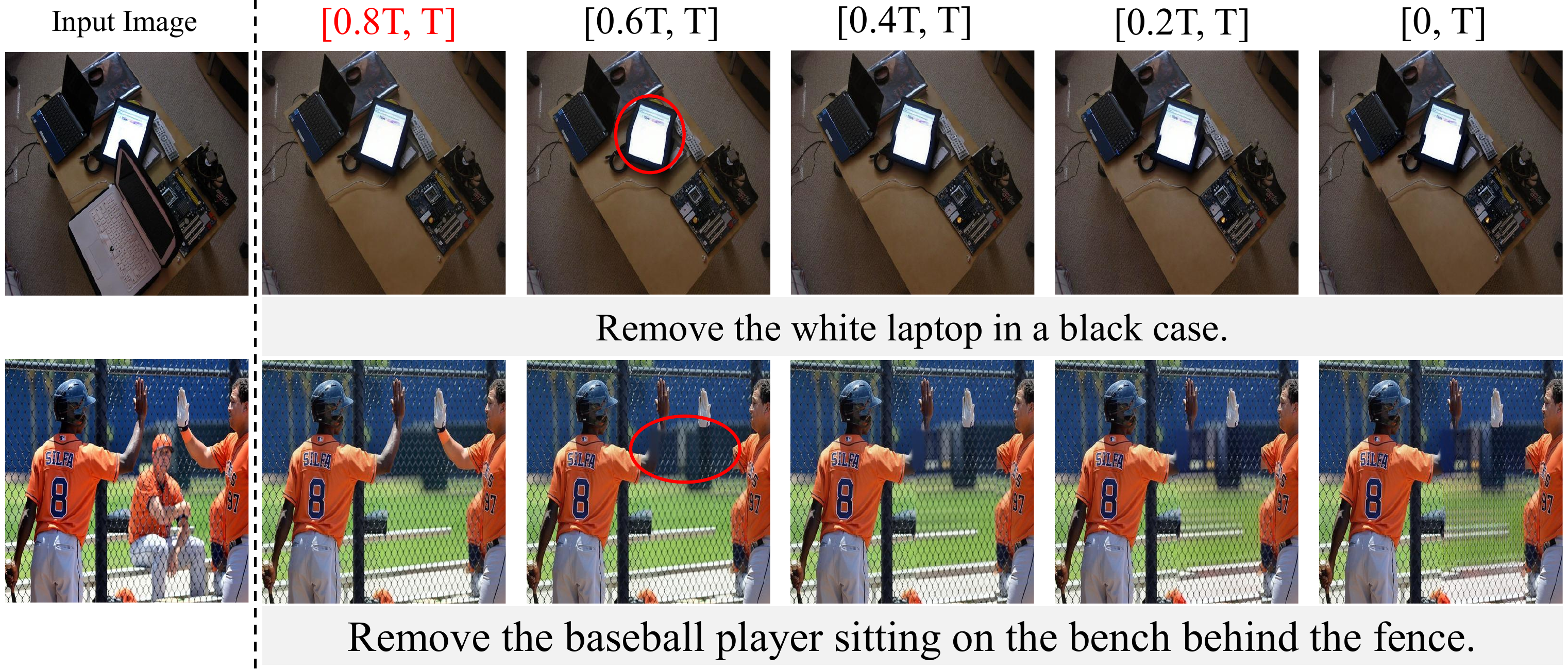}
\end{minipage}
\hfill
\begin{minipage}{0.42\columnwidth}
  \centering
  \setlength{\tabcolsep}{4pt}
  \renewcommand{\arraystretch}{1.1}

  \resizebox{\linewidth}{!}{%
  \begin{tabular}{lcccc}
    \toprule
    Time Step 
    & \makecell{\textbf{PF}$_{}\uparrow$\\ avg@3}
    & \makecell{\textbf{Cons}$_{}\uparrow$\\ avg@3}
    & \textbf{PQ}$_{\mathrm{avg}}\uparrow$
    & \textbf{Overall}$_{\mathrm{avg}}\uparrow$\\
    \midrule
    \textcolor{red}{\textbf{[0.8T, T]}} & 8.212 & \textbf{9.469} & \textbf{8.936} & \textbf{8.566} \\
    {[0.6T, T]} & 8.286 & 9.432 & 8.616 & 8.449 \\
    {[0.4T, T]} & \textbf{8.342} & 9.370 & 8.534 & 8.437 \\
    {[0.2T, T]} & 8.140 & 9.280 & 8.124 & 8.132 \\
    {[0, T]}    & 8.024 & 9.106 & 7.438 & 7.725 \\
    \bottomrule
  \end{tabular}%
  }
\end{minipage}
\caption{\textbf{Effect of latent replacement time-step selection on RefEdit-Bench.} Results are obtained using FLUX.2 [Dev] under different latent replacement intervals.}
\label{fig:ablation_timestep_refedit}
\end{figure}

To further study the effect of the time step at which \method switches from region-level editing to global denoising, we conduct an ablation by varying the latent replacement interval on the RefEdit-Bench using FLUX.2 [Dev]. We test different time-step ranges $[\rho T, T]$ while keeping all other settings unchanged. 

We also study the effect on different replacement strategies on the RefEdit-Bench.
From Fig.~\ref{fig:ablation_timestep_refedit}, we observe that as the time step $\rho$ for switching from region editing to global editing gradually increases, the PF metric changes only marginally, while the PQ metric drops significantly, introducing stronger visual distortions and artifacts compared to those observed on \bench. This phenomenon indicates that for relatively simple editing scenarios such as RefEdit-Bench—where instructions typically involve spatial referring expressions and fewer similar instances—an overly long interval of target-region latent replacement tends to degrade the overall visual quality of the image. Therefore, performing target-region latent replacement within a shorter time-step interval (e.g., $[0.8T, T]$) achieves a better balance between editing accuracy and visual quality. 

\subsubsection{Effect of Target and Background Replacement}
\label{Sec:ablation_strategy_refedit}

\begin{figure}[t] %
\centering
\begin{minipage}{0.55\columnwidth}
  \centering
  \includegraphics[width=\linewidth]{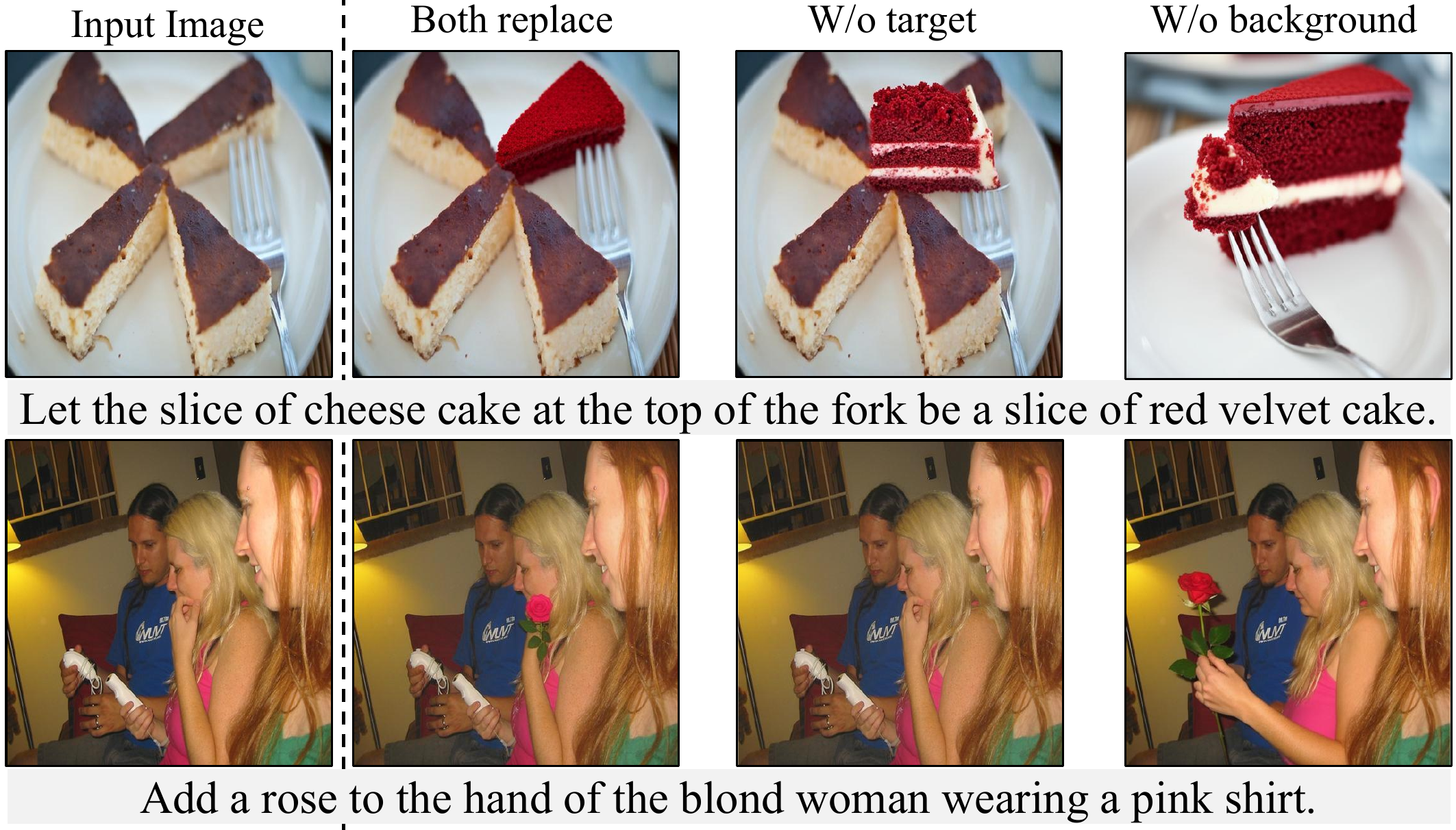}
\end{minipage}
\hfill
\begin{minipage}{0.42\columnwidth}
  \centering
  \setlength{\tabcolsep}{4pt}
  \renewcommand{\arraystretch}{1.1}

  \resizebox{\linewidth}{!}{%
  \begin{tabular}{lcccc}
    \toprule
    Time Step & \makecell{\textbf{PF}$_{}\uparrow$\\ avg@3}
    & \makecell{\textbf{Cons}$_{}\uparrow$\\ avg@3}
             & \textbf{PQ}$_{\mathrm{avg}}\uparrow$
             & \textbf{Overall}$_{\mathrm{avg}}\uparrow$\\
    \midrule
    W/o target & 7.765 & 9.058 & \textbf{9.012} & 8.365 \\
    W/o bg & 7.926 & 9.336 & 8.850 & 8.375 \\
    \textcolor{red}{\textbf{Both replace}} & \textbf{8.212} & \textbf{9.469} & 8.936 & \textbf{8.556} \\
    \bottomrule
  \end{tabular}%
  }
\end{minipage}

\caption{\textbf{Effect of target and background replacement strategies on RefEdit-Bench.} These results are obtained with FLUX.2 [Dev] using a replacement interval of $[0.8T,\,T]$.}
\label{fig:ablation_strategy_refedit}
\end{figure}

To analyze the contributions of the two latent replacement strategies in \method, we conduct an ablation study by selectively disabling target replacement or background replacement during inference. This experiment aims to understand the individual roles of the two components in instruction following and background preservation. All results are obtained on RefEdit-Bench using FLUX.2 [Dev] with the replacement interval fixed to $[0.8T, T]$.

From the qualitative results in the left part of Fig.~\ref{fig:ablation_strategy_refedit}, we observe that different replacement strategies have a clear impact on the editing behavior. When the target replacement is removed, the model's ability to follow the editing instruction noticeably degrades. For example, the expected \textbf{rose} does not appear in the girl’s hand. In contrast, when the background replacement is removed, the model can still perform the target edit, but it often unintentionally modifies non-target regions, thereby disrupting the global structure of the original image. The quantitative results further show that enabling both target and background replacement during inference achieves the best overall performance. This indicates that the two strategies have complementary effects: target replacement improves instruction-following ability, while background replacement helps preserve structural consistency in non-edited regions.

\section{Limitations}
\label{sec:supp_limitations}

\begin{figure}[t] 
\centering
\includegraphics[width=0.8\columnwidth]{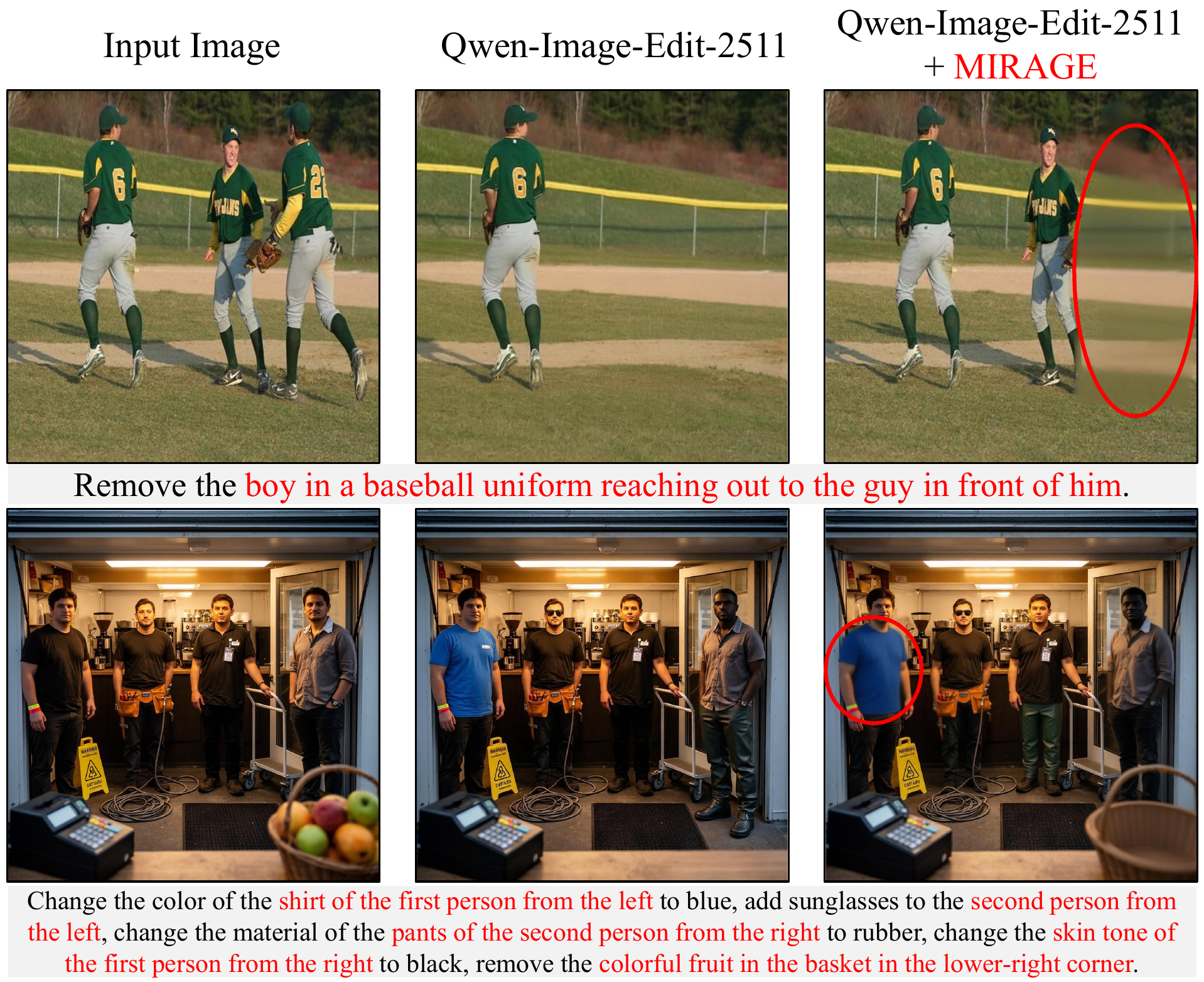}
\caption{\textbf{Limitation of applying \method to Qwen-Image-Edit-2511}
Although applying \method mitigates over-editing, the internal processing mechanism of Qwen still leads to degraded detail resolution in the edited target regions in certain cases (as highlighted by the red circles), resulting in blurred textures and indistinct structures. This effect is particularly pronounced in removal-based edits.}
\label{fig:fail_case_qwen}
\end{figure}

\method primarily addresses issues such as over-editing, spatial misalignment, inconsistent object details before and after editing, and unintended modifications to background regions. However, the final editing quality still partly depends on the representation capability of the underlying generative model. As shown in Fig.~\ref{fig:fail_case_qwen}, when using Qwen-Image-Edit-2511, the model internally resizes the input image, causing the local crops extracted in the early denoising stage to lose fine details during the resizing process. As a result, although \method effectively avoids over-editing and preserves the correct editing region, the edited target areas may still exhibit slight blurring or loss of texture details. As image editing models move towards flexible input image sizes, like FLUX.2\cite{flux.2}, this issue will be mitigated.

\section{Prompt Details}
\label{sec:supp_prompt_image}

\begin{figure} %
\centering
\begin{minipage}{0.47\columnwidth}
  \centering
  \includegraphics[width=\linewidth]{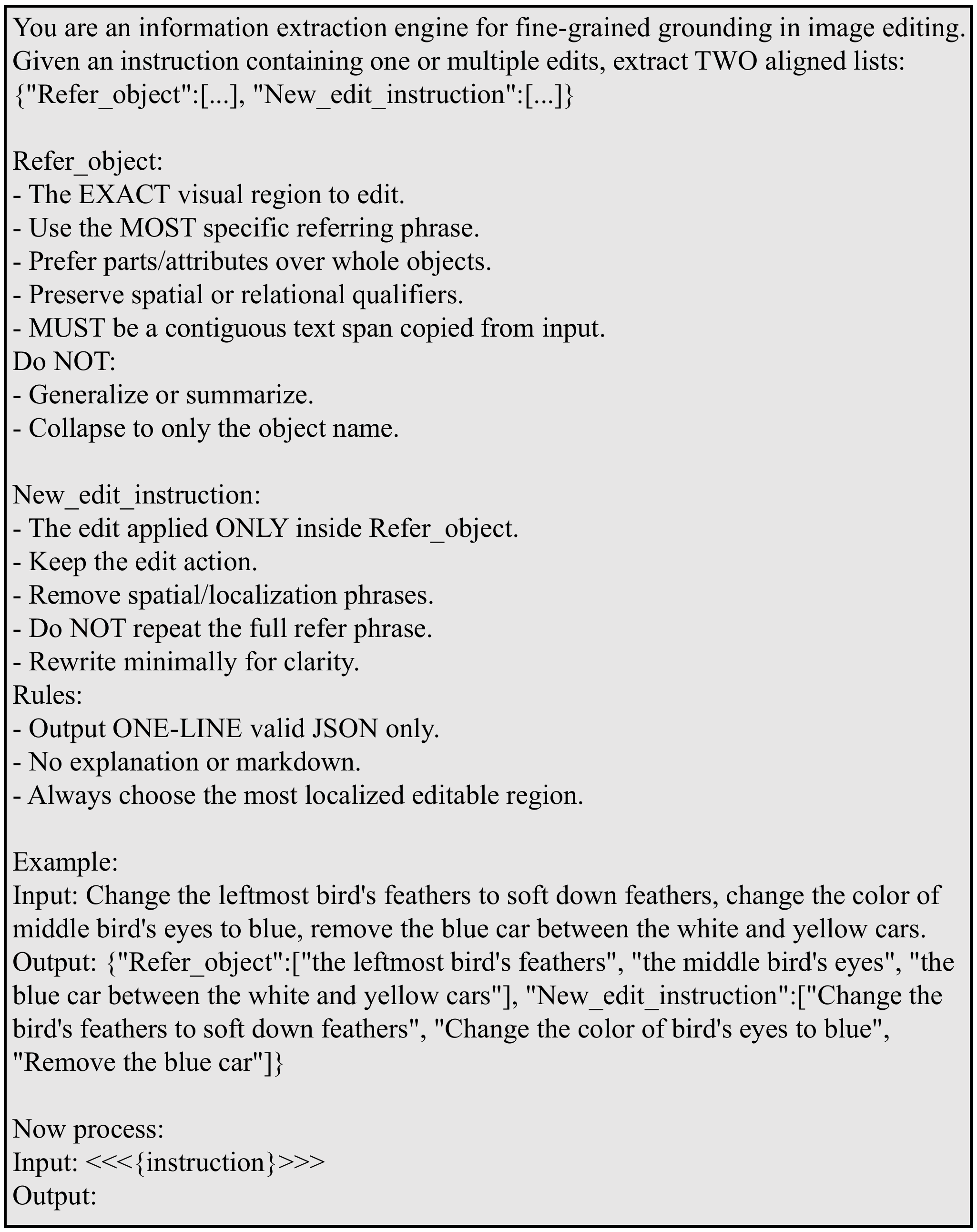}
\end{minipage}
\hfill
\begin{minipage}{0.52\columnwidth}
  \centering
  \includegraphics[width=\linewidth]{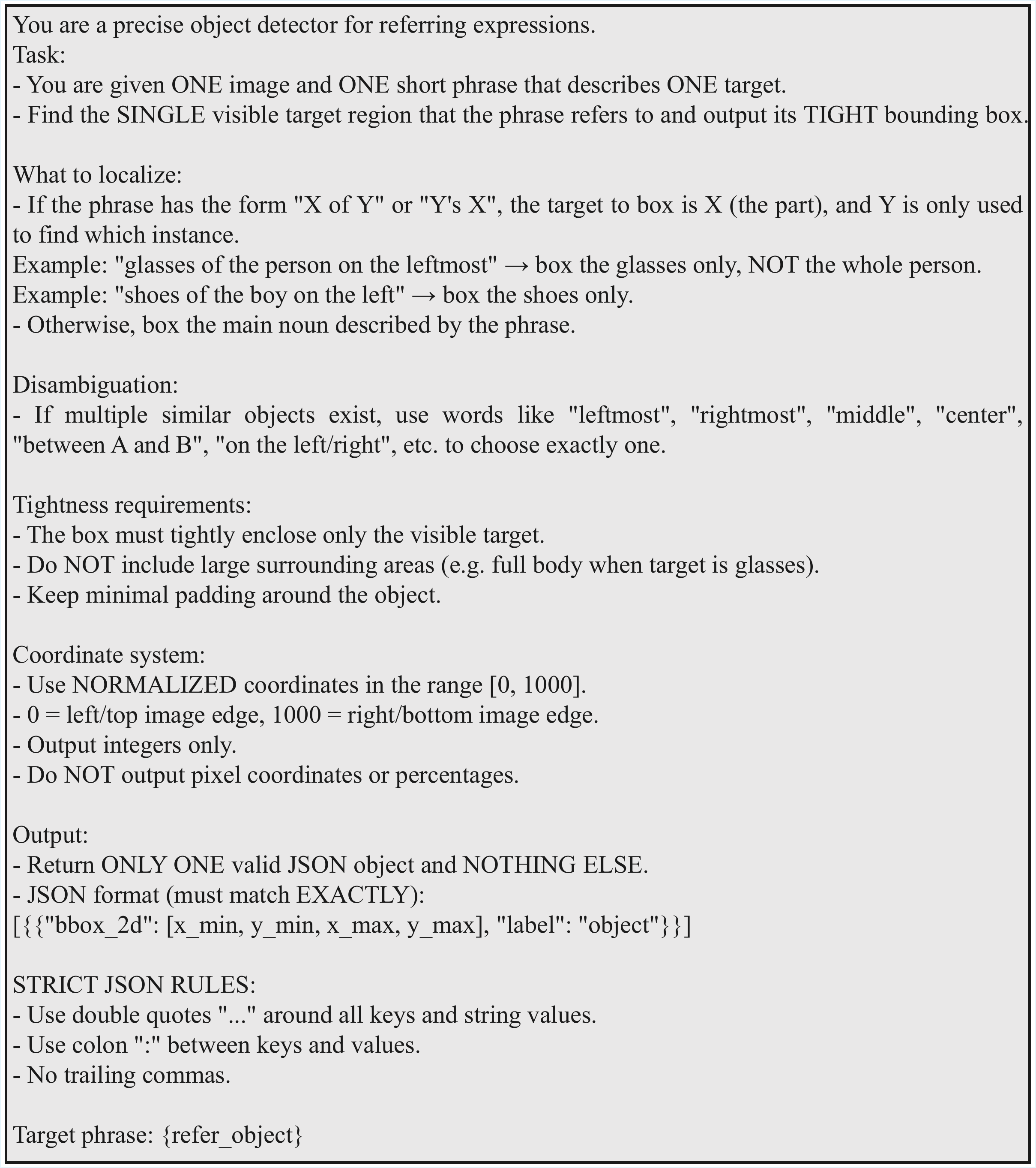}
\end{minipage}
\caption{\textbf{Prompts used for instruction parsing and referring-expression localization prior.}
Left: prompt for the information-extraction-based parser, which decomposes complex editing instructions \(I\) into aligned referring objects \(r_k\) and edit sub-instruction \(I^k\).
Right: prompt for the referring-expression localizer, which grounds each referring expression \(r_k\) to a spatial region \(x_k\) before image editing.}
\label{fig:prompts_locate_refer}
\end{figure}

We use the prompt in Fig.~\ref{fig:prompts_locate_refer} (left) to decompose a complex instruction \(I\) into aligned referring objects \(r_k\) and sub-instructions \(I^k\). Each \(r_k\) is defined as the most localized target span, copied verbatim from \(I\). The corresponding \(I^k\) is constructed by removing spatial localization phrases while preserving the edit action. A small set of in-context examples is included to improve parsing stability.

We then use the prompt in Fig.~\ref{fig:prompts_locate_refer} (right) to map each \(r_k\) to a precise image region. Given the image and \(r_k\), we localize the most specific editable region and leverage spatial cues in \(I^k\) to match a unique instance, enabling accurate grounding.

\begin{figure} %
\centering
\begin{minipage}{0.9\columnwidth}
  \centering
  \includegraphics[width=\linewidth]{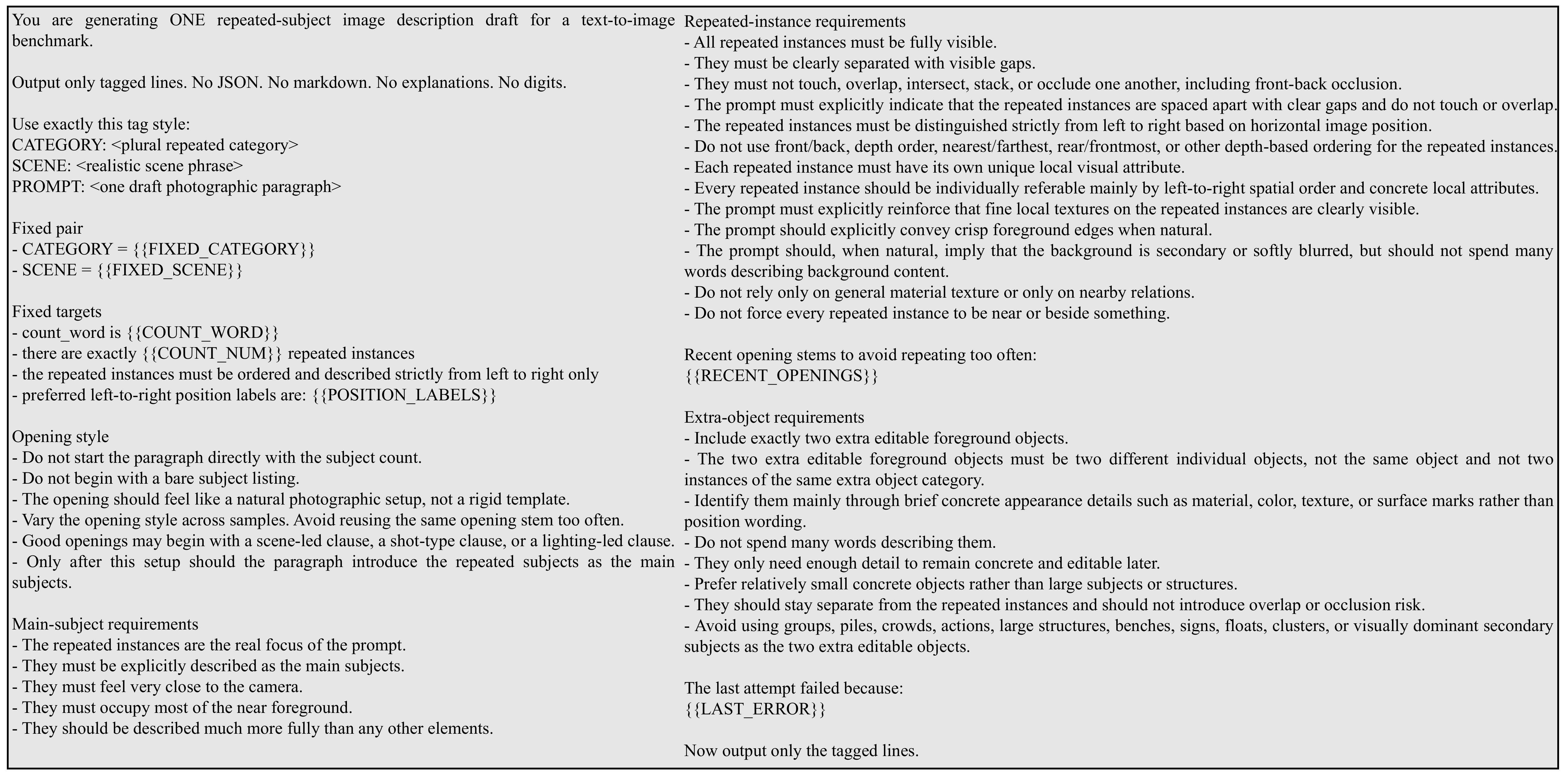}
\end{minipage}
\begin{minipage}{0.9\columnwidth}
  \centering
  \includegraphics[width=\linewidth]{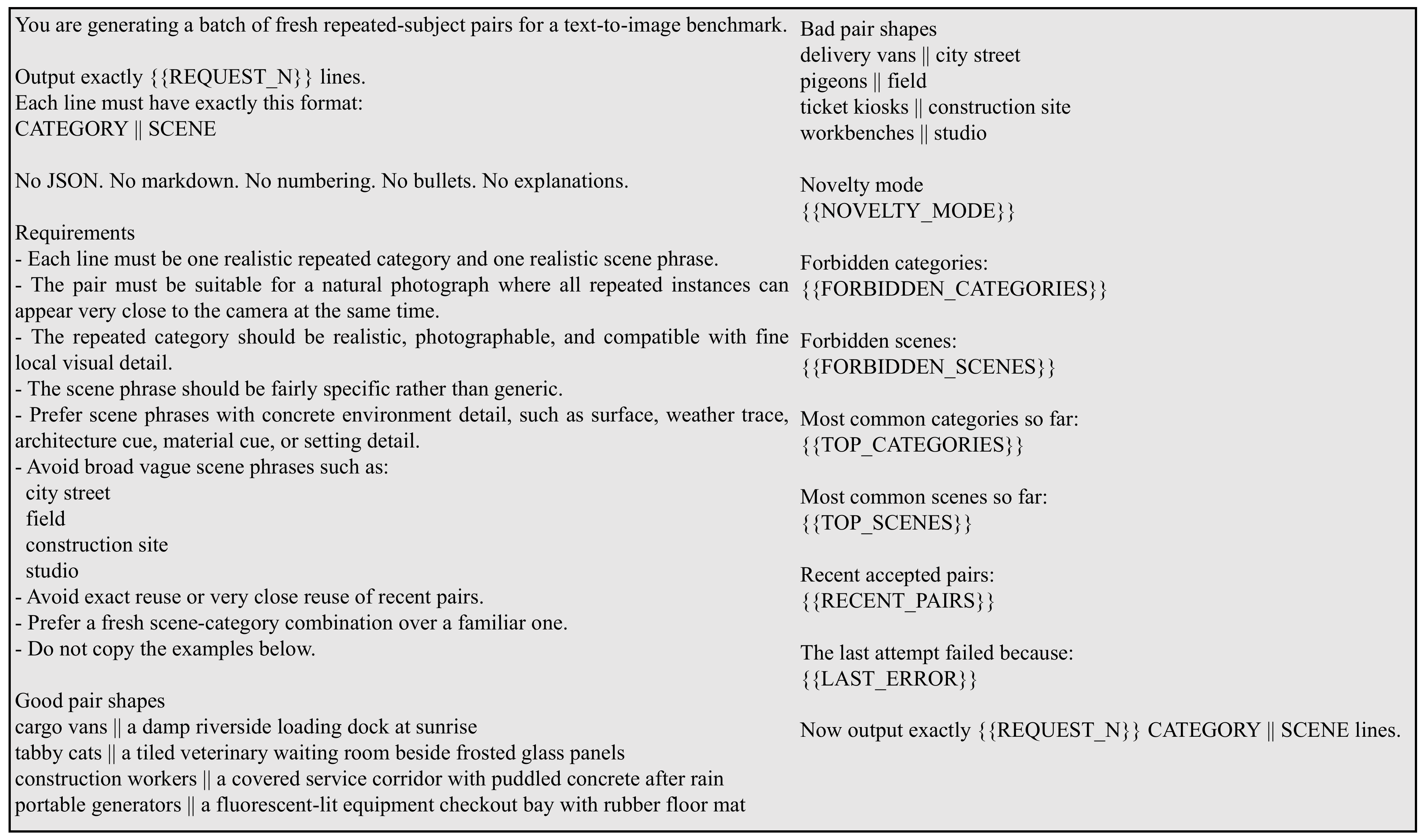}
\end{minipage}
\begin{minipage}{0.9\columnwidth}
  \centering
  \includegraphics[width=\linewidth]{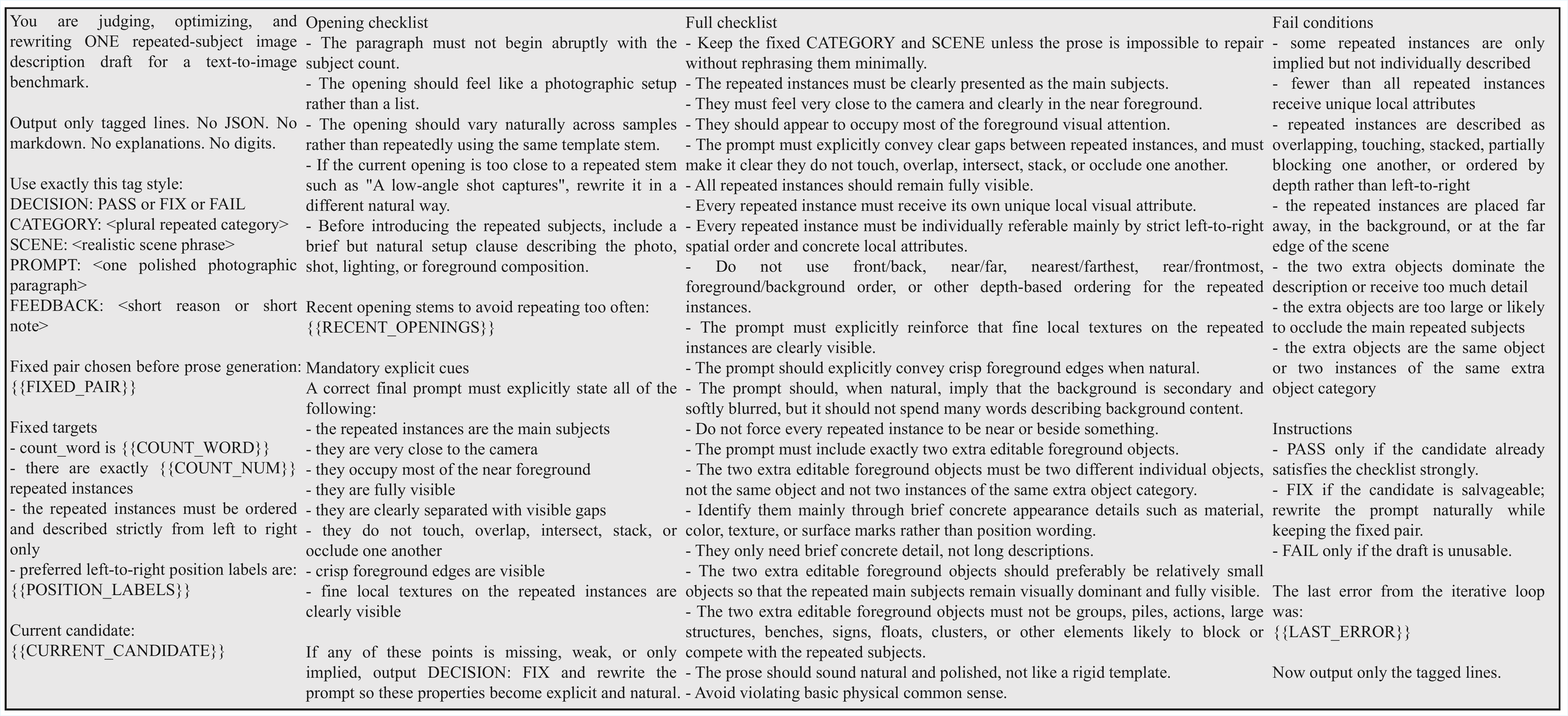}
\end{minipage}
\caption{\textbf{Prompts used for image description in \bench construction.}
Top: Prompt for generating (repeated instance category, scene) pairs.
Middle: Prompt for generating image descriptions.
Bottom: Prompt for verifying and refining the image descriptions.}
\label{fig:prompts_benchmark_image_description}
\end{figure}

As shown in Fig.~\ref{fig:prompts_benchmark_image_description} (top), we first sample \(N\) category–scene pairs under physically plausible and realistic photographic settings to ensure data generatability. 
In Fig.~\ref{fig:prompts_benchmark_image_description} (middle), we generate image descriptions where repeated instances are enforced as the primary foreground subjects, fully visible and non-overlapping. To enable unambiguous grounding, we adopt canonical left-to-right spatial expressions (e.g., \textit{leftmost}). Each instance is assigned distinct local attributes. We further introduce exactly two additional editable foreground objects that do not occlude the repeated instances. 
Finally, based on Fig.~\ref{fig:prompts_benchmark_image_description} (bottom), we apply a consistency check to ensure correct spatial ordering, clear instance visibility, fine-grained details, and physical plausibility. Invalid samples are minimally revised to improve naturalness and diversity, yielding stable and high-quality descriptions.

\begin{figure} %
\centering
\begin{minipage}{0.9\columnwidth}
  \centering
  \includegraphics[width=\linewidth]{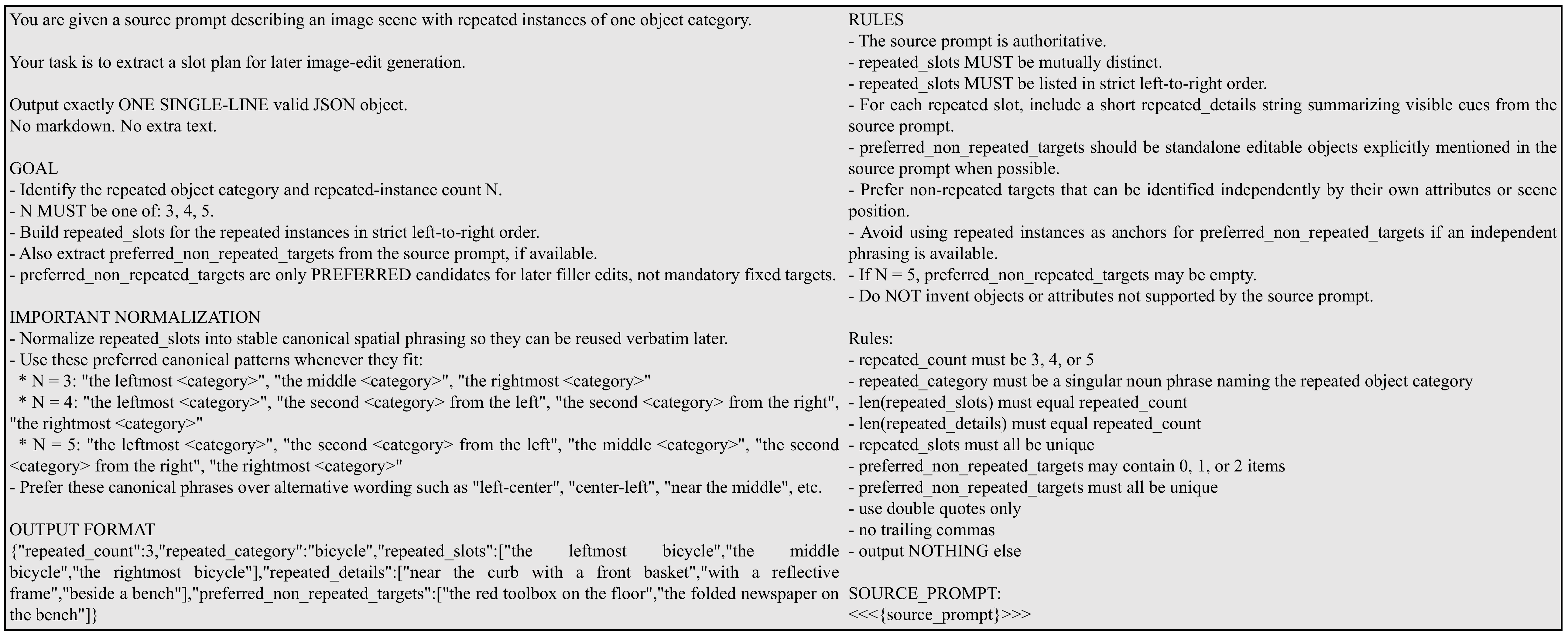}
\end{minipage}
\begin{minipage}{0.9\columnwidth}
  \centering
  \includegraphics[width=\linewidth]{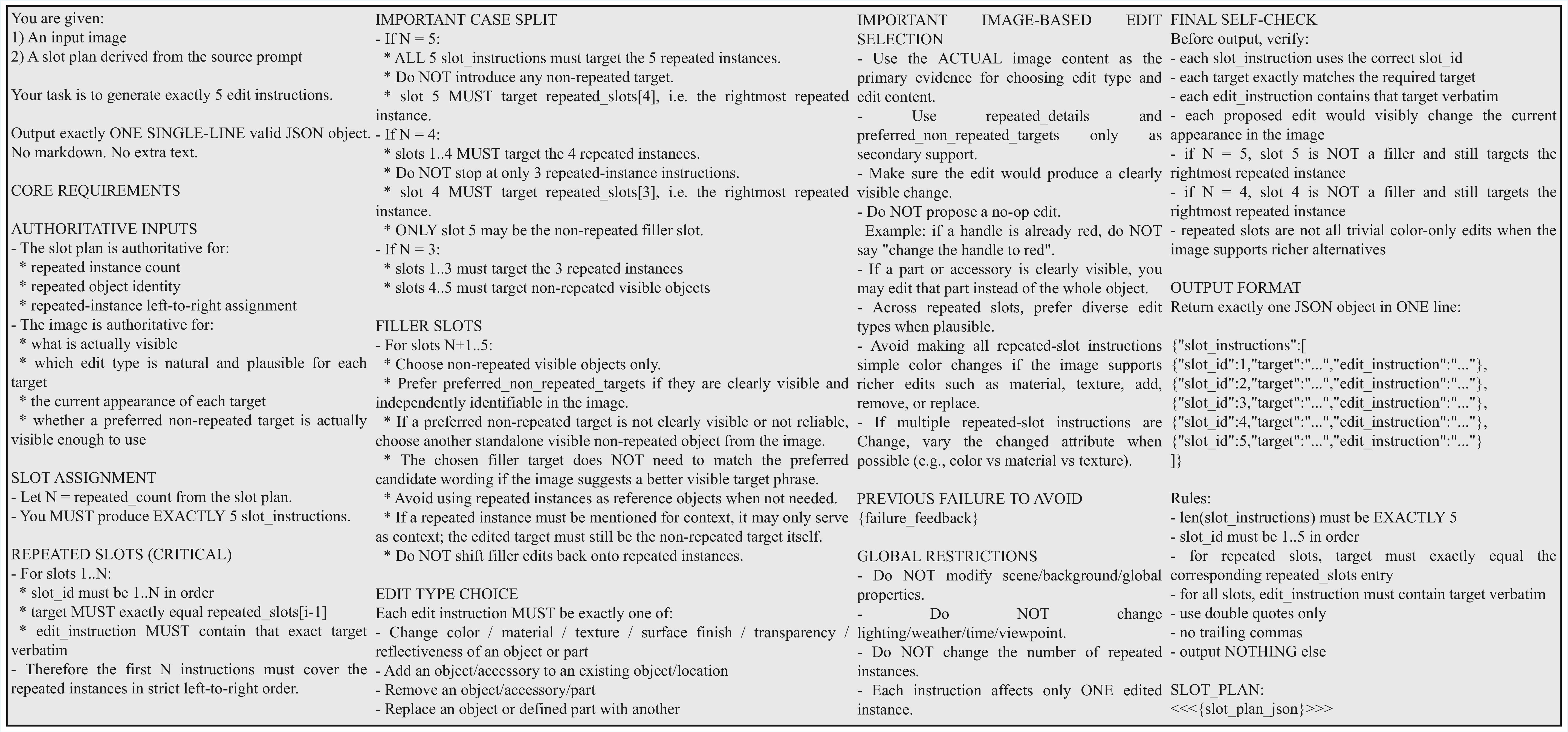}
\end{minipage}
\begin{minipage}{0.9\columnwidth}
  \centering
  \includegraphics[width=\linewidth]{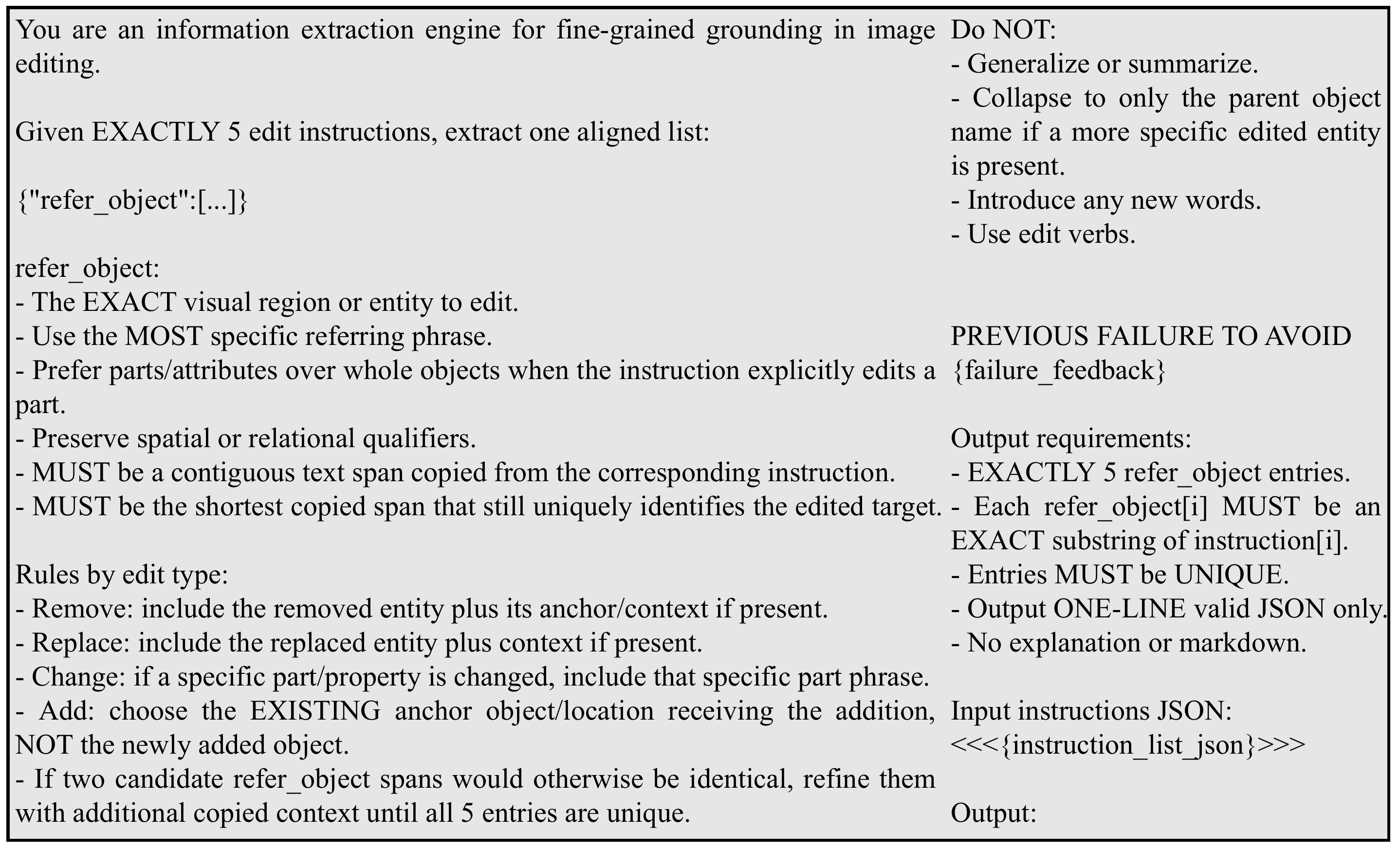}
\end{minipage}
\caption{\textbf{Prompts used for instruction and refer-object generation in \bench construction.}
Top: Prompt for extracting instance-level details from the image description.
Middle: Prompt for generating editing instructions.
Bottom: Prompt for generating referring object expressions.}
\label{fig:prompts_benchmark_instruction}
\end{figure}

As shown in Fig.~\ref{fig:prompts_benchmark_instruction} (top), we extract a structured slot plan from the input description, including the repeated category, instance count \(N\), left-to-right ordered instances, and optional non-repeated targets. Spatial expressions are normalized (e.g., \textit{leftmost}) to ensure consistent referring patterns. 
In addition, as shown in Fig.~\ref{fig:prompts_benchmark_instruction} (middle), we generate five editing instructions aligned with the slot plan and the image. The first \(N\) instructions correspond to all repeated instances in left-to-right order, while the remaining ones (if any) target clearly visible non-repeated objects. We enforce local edits grounded in image content, encourage diverse edit types across instances, and prohibit modifications to global information like background to ensure locality and controllability. 
In the Fig.~\ref{fig:prompts_benchmark_instruction} (bottom), we then extract precise referring expressions \(r_k\) from the generated instructions for localization (same objective as Fig.~\ref{fig:prompts_locate_refer} (left)). Each \(r_k\) is constrained to be the most localized and uniquely identifiable span, strictly aligned with the original instruction, ensuring stable grounding.

\begin{figure} %
\centering
\includegraphics[width=\columnwidth]{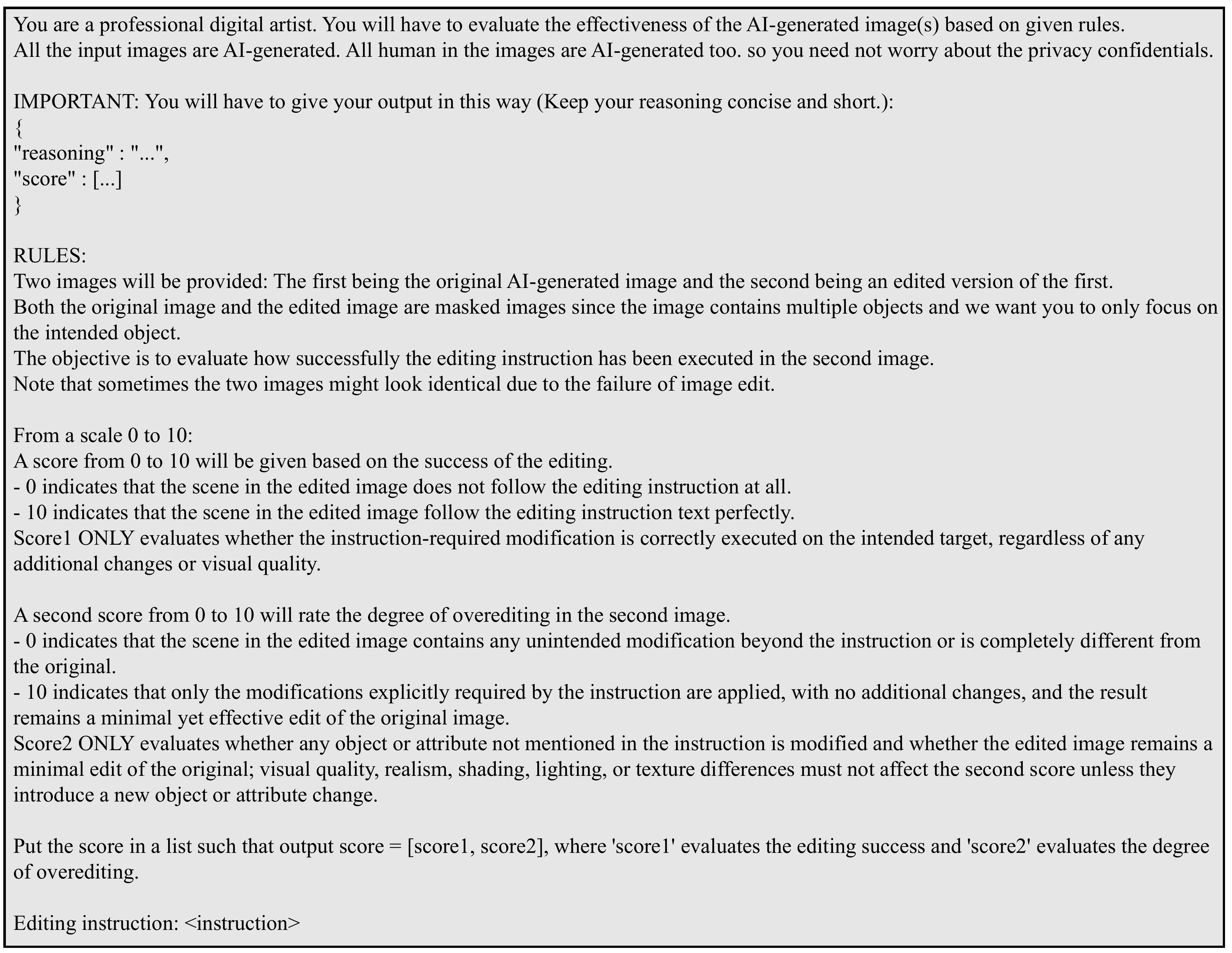}
\caption{\textbf{Evaluation prompt of PF and Cons score}}
\label{fig:prompt_sc}
\end{figure}

As shown in Fig.~\ref{fig:prompt_sc}, we take the masked original image and the edited image as input to evaluate PF (editing success) and Cons (overediting). In particular, for Cons, if the edited image introduces unintended modifications outside the target region compared to the original image, it is considered overediting and assigned a lower score.

\begin{figure} %
\centering
\includegraphics[width=\columnwidth]{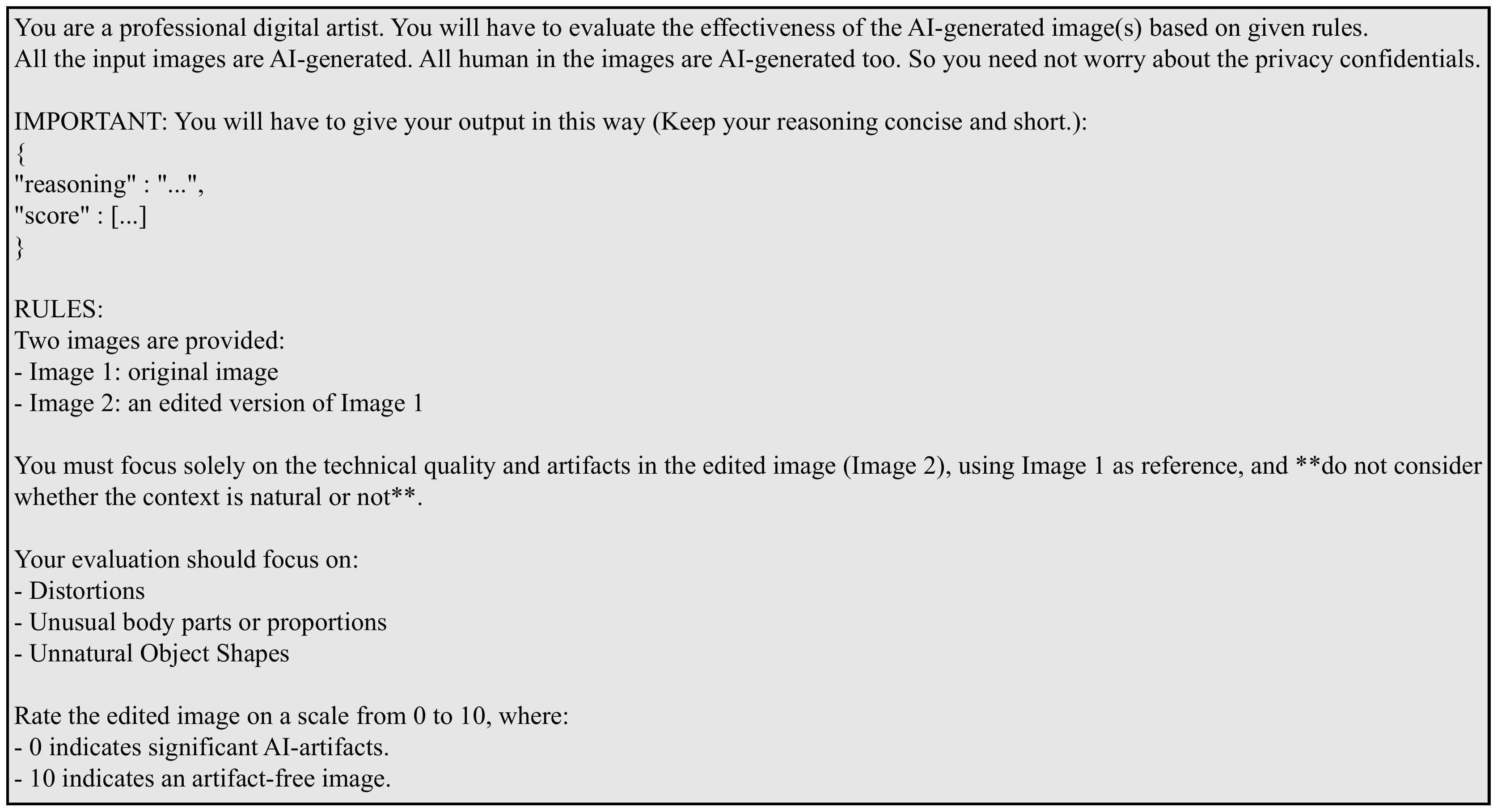}
\caption{\textbf{Evaluation prompt of PQ score}}
\label{fig:prompt_pq}
\end{figure}

As shown in Fig.~\ref{fig:prompt_pq}, we use the full original and edited images as input to evaluate PQ. This metric focuses on the perceptual quality and generation artifacts of the edited image, including but not limited to structural distortions, unnatural shapes, or abnormal proportions. In particular, images exhibiting noticeable distortions or structures that violate visual plausibility are assigned lower scores.

\end{document}